\documentclass[conference]{IEEEtran}
\usepackage{cite}
\usepackage{amsmath,amssymb,amsfonts}
\usepackage[noend]{algpseudocode}
\usepackage{textcomp}
\usepackage{xcolor}
\usepackage{lineno}

\usepackage[ruled,vlined,linesnumbered]{algorithm2e}

\usepackage{times}
\usepackage{latexsym}
\usepackage{subfig}
\usepackage{subcaption}
\usepackage{tabularx}
\usepackage{adjustbox}
\usepackage{booktabs}
\usepackage{pifont}
\usepackage{enumitem}
\usepackage[table]{xcolor}
\usepackage{threeparttable} 

\definecolor{shade}{RGB}{255,255,224}    

\usepackage{dsfont}
\usepackage{bbm}

\usepackage{adjustbox}
\usepackage{booktabs}
\usepackage{multirow}
\usepackage[ruled,linesnumbered,vlined]{algorithm2e}
\usepackage{makecell}
\usepackage{colortbl}
\definecolor{ourblue}{RGB}{219, 234, 254}  

\usepackage[most]{tcolorbox}

\newtcolorbox{takeawaybox}{
  colback=blue!5!white,
  colframe=blue!70!black,
  boxrule=1.2pt,
  arc=6pt,
  left=10pt,
  right=10pt,
  top=8pt,
  bottom=8pt
}

\newtcolorbox{questionbox}{
  colback=red!5!white,
  colframe=red!70!black,
  boxrule=1.2pt,
  arc=6pt,
  left=10pt,
  right=10pt,
  top=8pt,
  bottom=8pt
}

\definecolor{llamagreen}{RGB}{198,239,206}
\definecolor{ligerbase}{RGB}{255,218,224}
\definecolor{quantblue}{RGB}{173,216,230}
\usepackage[T1]{fontenc}

\usepackage[utf8]{inputenc}

\usepackage{microtype}

\usepackage{inconsolata}

\usepackage{tikz}

\usepackage{graphicx}

\def\BibTeX{{\rm B\kern-.05em{\sc i\kern-.025em b}\kern-.08em
    T\kern-.1667em\lower.7ex\hbox{E}\kern-.125emX}}

\usepackage{url}
\usepackage{hyperref}

\begin{document}

\setcounter{page}{0}

\title{GLIDE: Guided Layerwise Hybrid Attention for Efficient LLM Inference\\
}


\author{
    \IEEEauthorblockN{Vimal William \quad Ravi Tandon \quad Jyotikrishna Dass}
    \IEEEauthorblockA{\textit{Department of Electrical \& Computer Engineering, University of Arizona, Tucson, USA}}
    \IEEEauthorblockA{Email: \{\texttt{vimalwilliam}, \texttt{tandonr}, \texttt{jdass}\}@arizona.edu}
}

\maketitle

\begin{abstract}

As Large Language Models scale to increasingly long contexts, the memory I/O and computational overhead of the Key-Value (KV) cache during decoding emerges as the primary throughput bottleneck. To address this, we propose \textbf{\textsc{Glide}}, a Guided Layerwise Hybrid Attention that strategically integrates sliding-window softmax attention with linear recurrent aggregation. \textbf{\textsc{Glide}} is motivated by layer-wise heterogeneity: early layers exhibit high sensitivity to softmax removal, while deeper layers demonstrate redundancy and tolerate aggressive replacement by linear alternatives. Leveraging this insight, \textbf{\textsc{Glide}} introduces a \textit{layer-wise adaptive mechanism} wherein each layer balances an efficient linear recurrence with a variable-sized softmax window. Unlike uniform hybrid approaches, \textbf{\textsc{Glide}} non-uniformly compresses the softmax footprint across the model, reducing \textit{aggregate KV cache I/O} while preserving expressive power where most vital. Empirical evaluations demonstrate the \textbf{\textsc{Glide}} achieves superior performance-efficiency tradeoffs, reducing end-to-end latency for long-context generation without compromising quality.

\end{abstract}


\section{Introduction}
Long-context Large Language Model (LLM) inference is increasingly limited not by compute alone, but by data movement during autoregressive decoding. As the KV-cache grows with sequence length, each generated token demands retrieval of an expanding volume of cached key-value states, shifting execution from compute-bound to memory-bandwidth-bound regimes~\cite{feng2024ada}. This bottleneck is particularly acute in long-context generation, where KV-cache growth inflates both memory footprint and off-chip data transfer, degrading hardware utilization and overall throughput. Such constraints pose a critical barrier to scalable deployment on memory-constrained accelerators and in latency-sensitive serving environments~\cite{li2025a}. This motivates a central question: \textit{How can long-context decoding achieve lower latency and reduced memory I/O while maintaining generation quality?}


\vspace{1mm}
\noindent Recent work has approached efficient long-context inference by optimizing KV-cache management through two main strategies: a) \textit{Eviction-based} methods reduce memory and compute by selectively discarding cached tokens deemed less relevant~\cite{zhang2023h2o, cai2024pyramidkv, li2024snapkv}, while b) \textit{Retention-based} methods preserve contextual information through hybrid attention mechanisms that combine local softmax attention with efficient recurrent-style aggregation. Retention-based hybrid architectures  (also see Section~\ref{sec: perf} such as LoLCats~\cite{zhang2025lolcats} and Liger~\cite{lan2025liger}) offer a particularly promising design point: they apply parameter-efficient fine-tuning (PEFT) to distill pretrained softmax attention into linear recurrent forms, improving decoding efficiency without explicitly evicting context from the KV-cache. However, existing retention-based methods apply a uniform hybridization policy across all layers, replacing softmax attention with linear alternatives at fixed positions throughout the network. This one-size-fits-all strategy overlooks a critical observation: attention layers vary significantly in their sensitivity to linearization. Early layers with foundational token representations may rely more heavily on precise softmax attention, while deeper layers operating on increasingly abstract features may tolerate aggressive approximation with minimal performance degradation. 


\begin{questionbox}
\textit{Can a non-uniform, layer-wise hybrid attention policy guided by depth-dependent sensitivity achieve a superior efficiency–accuracy Pareto frontier compared to uniform hybridization?}
\end{questionbox}

 \begin{figure*}[t]
    \centering
    \subfloat[\textbf{Various Attention Mechanisms}]{%
        \raisebox{-0.5mm}{\includegraphics[width=0.50\textwidth]{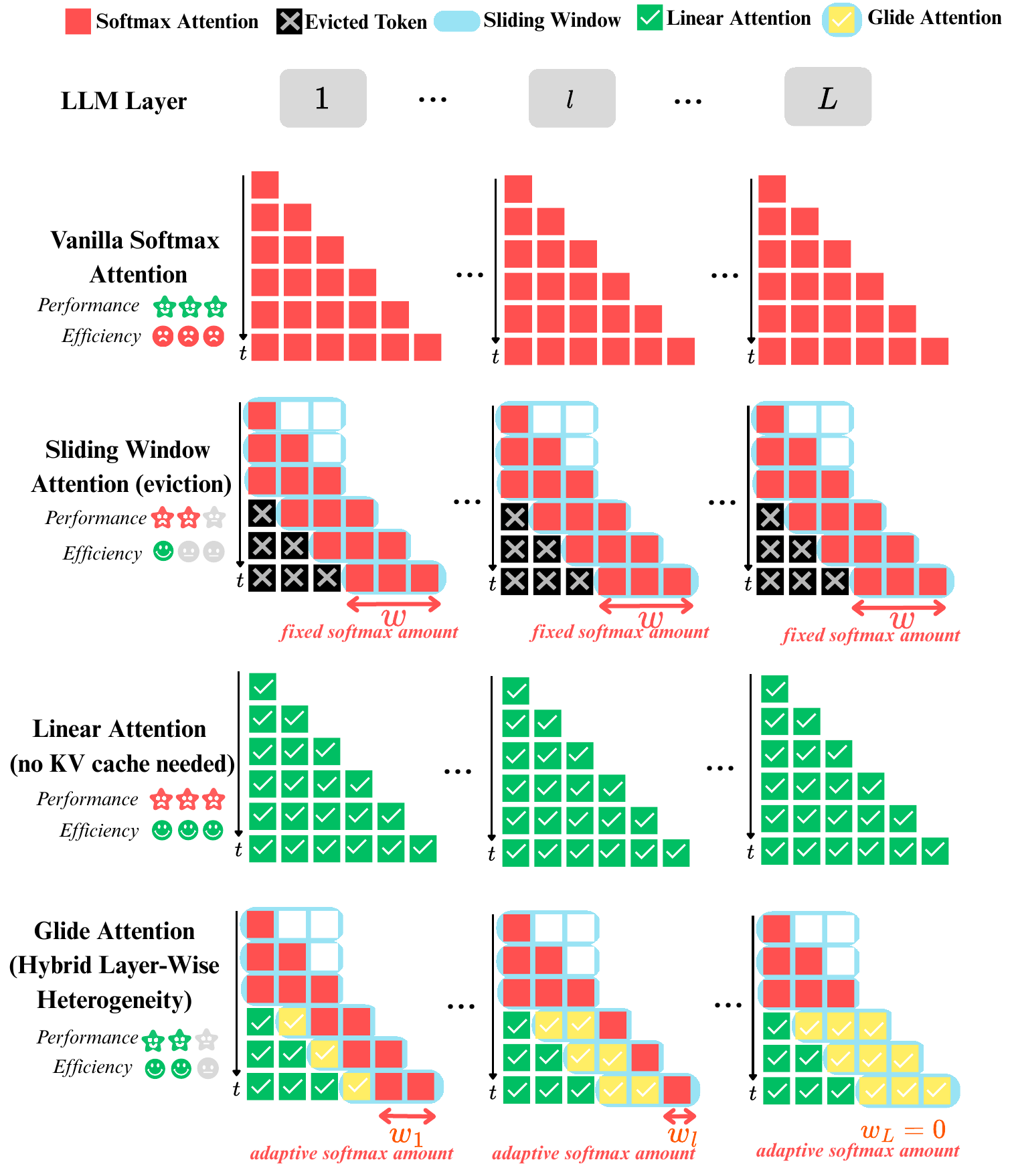}}%
        \label{fig:various_attention}%
    }%
    \hfill
    \subfloat[\textbf{Performance-Efficiency Pareto Analysis}]{%
        \includegraphics[width=0.50\textwidth]{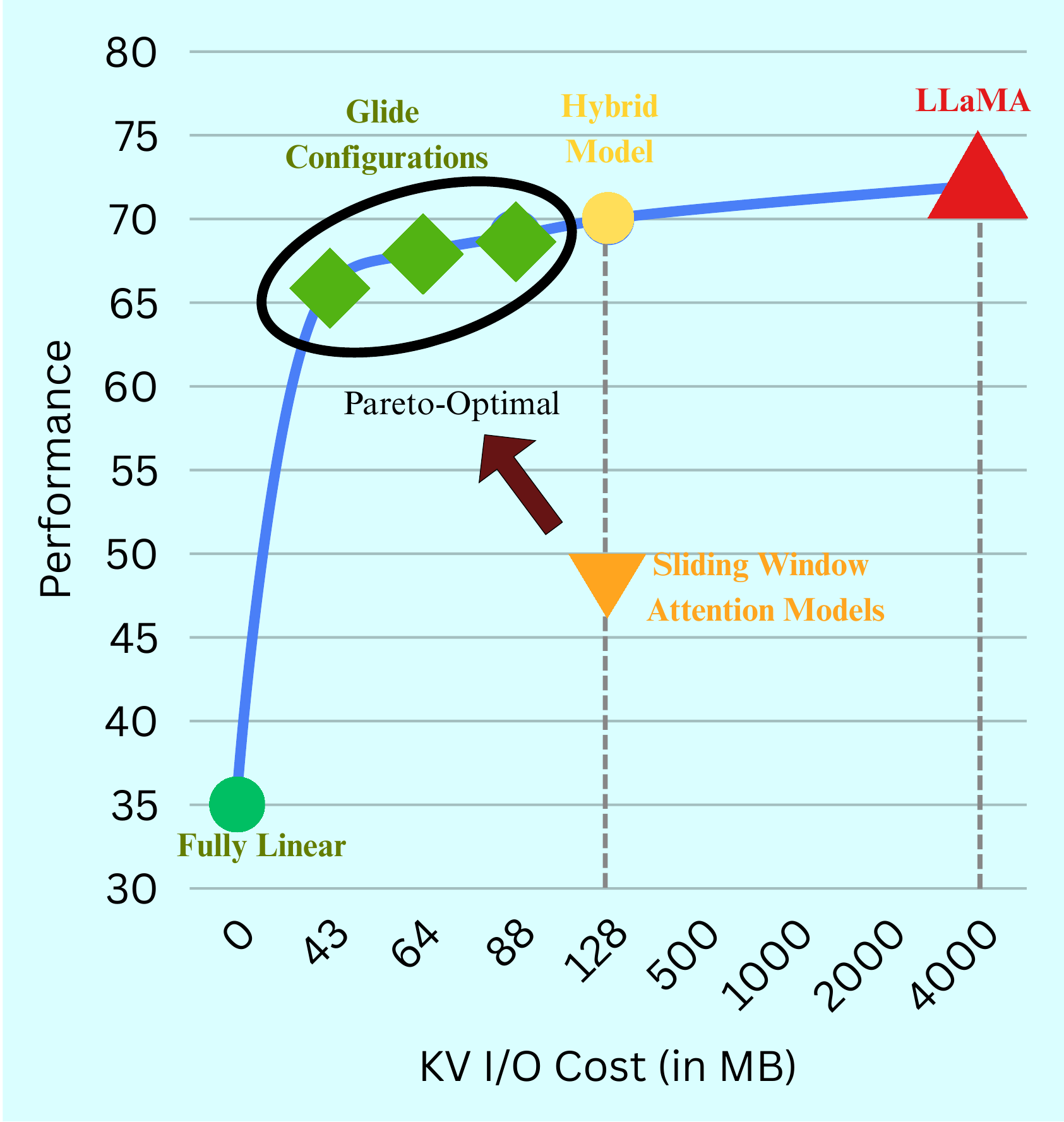}%
        \label{fig:pareto}%
    }
    \caption{\textbf{Layer-Wise Adaptive Attention Strategy of \textbf{\textsc{Glide}}.} \textbf{(a) Attention Mechanisms:} \textbf{\textsc{Glide}} strategically transitions from softmax attention (red) in early layers to linear attention (green) in deeper layers, concentrating expensive computation where most critical, unlike vanilla softmax, sliding window attention (SWA), or pure linear approaches. \textbf{(b) Pareto Analysis:} \textbf{\textsc{Glide}} configurations achieve $45\times$ -- $62\times$ lower KV cache I/O than baseline LLaMA while retaining $92\%$ -- $96\%$ accuracy, outperforming uniform alternatives (SWA, Hybrid Model) by exploiting layer-wise heterogeneity.}
    \label{fig:hero}
\end{figure*}

\vspace{1mm}

As a first step towards answering this question, we conduct an empirical study in which we systematically vary the ratio of softmax-to-linear attention across transformer layers and measure the resulting impact on downstream task accuracy. Our analysis reveals a clear layerwise heterogeneity in softmax importance: earlier layers responsible for establishing low-level token representations and local syntactic structure benefit significantly from retaining softmax-based attention, whereas deeper layers, which operate on more abstract semantic features, tolerate aggressive linearization with minimal impact on generation quality. 
This finding suggests that uniform softmax allocation across layers is fundamentally suboptimal, leaving substantial efficiency gains unrealized. Building on this insight, we introduce \textbf{\textsc{Glide}} (\textbf{G}uided \textbf{L}ayerwise Hybr\textbf{i}d Attention for Efficient \textbf{D}ecoding), a retention-based architecture that allocates softmax attention non-uniformly across layers according to their measured sensitivity.

\vspace{1mm}


\noindent As illustrated in Fig.~\ref{fig:hero}, \textbf{\textsc{Glide}} exploits layer-wise sensitivity patterns to achieve superior efficiency-accuracy trade-offs compared to various attention schemes. Fig.~\ref{fig:hero}(a) contrasts four attention mechanisms: vanilla softmax attention (red) applies full quadratic attention uniformly across all layers, achieving optimal performance but incurring prohibitive memory costs; sliding window attention (red with evicted tokens) reduces memory by discarding distant context, sacrificing accuracy; linear attention (green) eliminates KV cache overhead entirely but suffers severe degradation; and \textbf{\textsc{Glide}} strategically transitions from softmax attention in early layers to linear attention in deeper layers, concentrating expensive computation where most critical. This layer-wise allocation directly reduces KV cache memory traffic during autoregressive decoding, the dominant bottleneck in long-context generation, while preserving representational fidelity in sensitive layers. Fig.~\ref{fig:hero}(b) demonstrates the resulting Pareto frontier: \textbf{\textsc{Glide}} configurations occupy the favorable top-left region, achieving $45\times$ -- $62\times$ lower KV cache I/O than baseline while retaining $92\%$ -- $96\%$ accuracy, substantially outperforming uniform alternatives including hybrid models and sliding window attention which fail to exploit layer-wise heterogeneity. By enabling strategic allocation of softmax versus linear attention across depth, \textbf{\textsc{Glide}} unlocks previously inaccessible operating points that balance memory bandwidth, latency, and task performance. Our main contributions are as follows:

\begin{itemize}
    \item We conduct a systematic study of hybrid attention sensitivity across transformer depth, revealing that softmax attention importance is strongly layer-dependent: early layers are markedly more sensitive to linearization, with full linearization causing catastrophic accuracy collapse (36\% avg.), while late layers tolerate complete linearization with minimal degradation (Section~\ref{sec: perf}).

    \item We introduce \textbf{\textsc{Glide}}, a guided layerwise hybrid attention framework that allocates softmax attention non-uniformly across transformer blocks based on observed layer-wise sensitivity. \textsc{Glide} partitions the network into early, middle, and late blocks, assigning full softmax ($\delta = 0$) to early layers, partial linearization ($\delta = \alpha \cdot w$, $\alpha \in [0,1]$) to middle layers, and full linearization ($\delta = w$) to late layers. This reduces the $\delta$-configuration search space to a single scalar $\alpha$ that continuously interpolates between full softmax and full linearization (Section~\ref{sec:glide_method}).
    
    \item We demonstrate that \textbf{\textsc{Glide}} integrates seamlessly with existing retention-based architectures (Liger, LoLCats), achieving up to \textbf{62$\times$} reduction in KV-cache I/O and \textbf{3.3$\times$} decoding speedup while retaining 92--94\% of baseline accuracy across six reasoning benchmarks on Llama-3-8B and Mistral-7B (Section~\ref{sec:glide_exp}).
\end{itemize}

\section{Background}
This section gives an overview of attention and KV-cache mechanisms in autoregressive Transformer inference, reviews prior approaches for improving long-context efficiency, and highlights the gaps that motivates \textbf{\textsc{Glide}}.

\subsection{Notations}

We consider an $L$-layer Transformer model with $H$ attention heads per layer. Let $d_{\text{model}}$ denote the model dimension and $d = d_{\text{model}}/H$ denotes the per-head dimension. For autoregressive generation at position $i$, we denote the query vector as $\mathbf{q}_i \in \mathbb{R}^d$ and the key-value pairs as $\{\mathbf{k}_j, \mathbf{v}_j\}_{j=1}^{i}$ with $\mathbf{k}_j, \mathbf{v}_j \in \mathbb{R}^d$. The attention weights are denoted $\alpha_{ij}$, representing the normalized similarity between query $\mathbf{q}_i$ and key $\mathbf{k}_j$. The attention output at position $i$ is written as $\mathbf{O}_i$, with superscripts distinguishing variants: $\mathbf{O}_i^{\text{SWA}}$ for sliding window attention, $\mathbf{O}_i^{\text{Linear}}$ for linear attention, and $\mathbf{O}_i^{\text{Glide}}$ for \textbf{\textsc{Glide}} attention. We use $w$ to denote the sliding window size and $\delta$ to control the extent of linearization within the window, where $\boldsymbol{\delta_\ell}$ specifies the layer-wise allocation for layer $\ell \in [1, L]$. In linear attention, $\phi(\cdot)$ denotes the separable kernel (feature map) applied to queries and keys, while $\mathbf{S}_i \in \mathbb{R}^{d \times d}$ and $\mathbf{Z}_i \in \mathbb{R}^d$ represent the recurrent state and normalization factor, respectively. The operator $\oplus$ denotes the weighted combination of linear and softmax attention outputs based on their respective normalizing denominators. For block-wise allocation, the model is partitioned into early, middle, and late segments containing $L_e$, $L_m$, and $L_l$ layers, and the configuration tuple $\boldsymbol{\delta}_b = (\delta_{b_1}, \delta_{b_2}, \delta_{b_3})$ specifies the $\delta$ values assigned to each segment, where the $\delta = w.\alpha$ is generic definition of $\delta$ and we distinguish the attention score $\mathbf{\alpha_{i}} \in \mathbb{R}^{n}$ from the cache sparsity factor $\alpha \in [0, 1]$. 

\subsection{Preliminaries on Attention and KV Cache}

\textbf{Overview of Self-Attention.}
In autoregressive next-token prediction at position $i$, the query vector $\mathbf{q}_i \in \mathbb{R}^{d}$ (where $d = d_{\text{model}}/H$ for $H$ attention heads) attends to all previously generated keys $\{\mathbf{k}_j\}_{j=1}^{i}$ via scaled dot-product similarity, producing softmax-normalized attention weights that quantify relevance. The attention output $O_i$ is then obtained as the weighted aggregation of the corresponding value vectors $\{\mathbf{v}_j\}_{j=1}^{i}$ and subsequently linearly projected through $W_o$ to form the final head contribution. \textit{Consequently, predicting the next token requires access to the entire set of prior key–value pairs $\{ \mathbf{k}_j, \mathbf{v}_j \mid j \le i \}$, resulting in a quadratic computational complexity with respect to the sequence length during full attention computation.}
\vspace{1mm}


\textbf{Linearized Attention.} Linear attention~\cite{katharopoulos2020transformers} approximates softmax attention by replacing the exponential similarity kernel with separable feature mappings $\phi(\cdot)$ applied independently to queries and keys: $\alpha_{ij} \approx \frac{\phi(\mathbf{q}_i)\phi(\mathbf{k}_j)^{\top}}{\sum_{j=1}^{i}\phi(\mathbf{q}_i)\phi(\mathbf{k}_j)^{\top}}$. This enables constant-space recurrent computation: $O_i^{\text{Linear}} = \frac{\phi(\mathbf{q}_i)\mathbf{S}_i}{\phi(\mathbf{q}_i)\mathbf{Z}_i}$, where cumulative states $\mathbf{S}_i = \mathbf{S}_{i-1} + \phi(\mathbf{k}_i)^{\top}\mathbf{v}_i$ and $\mathbf{Z}_i = \mathbf{Z}_{i-1} + \phi(\mathbf{k}_i)^{\top}$ accumulate key-value interactions. Memory complexity is reduced from $O(n)$ to $O(1)$ with respect to context length, and KV cache I/O is eliminated entirely. \textit{However, kernel-based linear attention sacrifices expressiveness~\cite{zhang2402hedgehog, bridging_han_2024}: the separable approximation degrades attention quality, particularly in early transformer layers where precise token interactions are critical.}

\vspace{1mm}
\textbf{KV Cache.}
To avoid recomputing past states during autoregressive inference, KV caching stores generated key–value pairs $(\mathbf{k}_i, \mathbf{v}_i)$ for reuse. The process involves a \textit{prefill} stage that initializes the cache from the prompt, followed by a \textit{decode} stage where tokens are generated sequentially by attending to cached entries and appending new pairs. \textit{Although this removes redundant computation, the cache grows linearly with context length $n$, and each decoding step requires memory access, which grows linearly with respective to $n$}. 

 \begin{figure*}[t]
    \centering
    \subfloat[\textbf{GLIDE Configuration}]{%
        \raisebox{-0.5mm}{\includegraphics[width=0.25\textwidth]{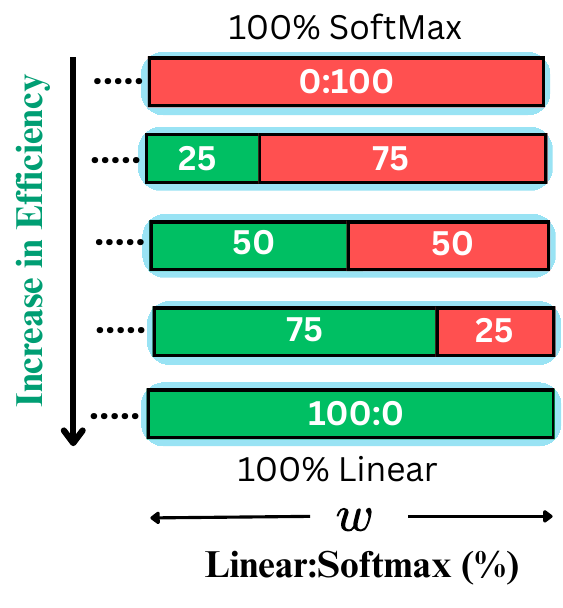}}%
        \label{fig:delta_ratio}%
    }%
    \hfill
    \subfloat[\textbf{All Layers/Isolated Sensitivity Pair-wise Analysis}]{%
        \includegraphics[width=0.74\textwidth]{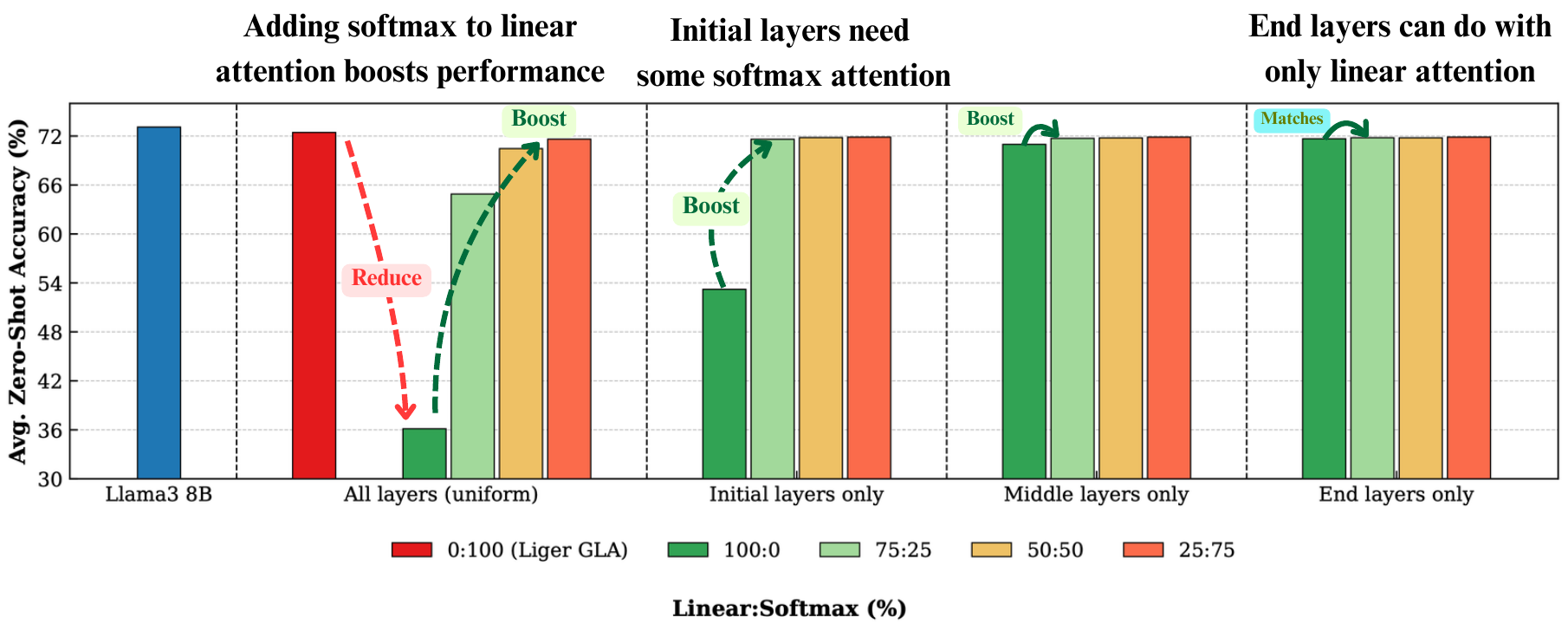}%
        \label{fig:layer_sensitivity}%
    }
    \caption{\textbf{Analysis of Zero-Shot Accuracy Under Varying Softmax Attention Intensity (Non~Fine-Tuned).} (a) The $\delta$ parameter controls the Linear:Softmax ratio within each layer's sliding window $w$, trading efficiency for representational fidelity. (b) Layer-wise sensitivity analysis across Early, Middle, and Late layer groups, evaluated on LM-Eval benchmarks (PiQA, ARC-e, ARC-c, HellaSwag, WinoG) with $w=64$ and $d=4096$. GLA (Gated Linear Attention) replaces softmax within the window using a fixed-size recurrent state, preserving context beyond the sliding window; the red bar ($\delta = w$) denotes full linearization.}
    \label{fig:abs_1}
\end{figure*}

\subsection{Policies of Sliding Window Attention}
\textbf{Sliding Window Attention.} To mitigate the linear memory growth and per-token attention cost of full KV cache, prior work proposes Sliding Window Attention (SWA) \cite{beltagy2020longformer}, where the query $\mathbf{q}_i$ attends only to the most recent $w$ key–value pairs. Concretely, attention weights are computed over the restricted range $j \in [i-w+1, i]$, i.e., $\alpha_{ij}^{\text{SWA}} = \frac{\exp(\mathbf{q}_i \mathbf{k}_j^\top)}{\sum_{j'=i-w+1}^{i} \exp(\mathbf{q}_i \mathbf{k}_{j'}^\top)}$, and the output is $O_i^{\text{SWA}} = \sum_{j=i-w+1}^{i} \alpha_{ij}^{\text{SWA}} \mathbf{v}_j$. This reduces the per-token memory and compute complexity to $O(w)$ with $w < n$. \textit{However, SWA follows a first-in-first-out eviction policy, whereby older tokens are discarded as new ones enter the window, potentially removing semantically important context and degrading recall over long-range dependencies.} We categorize prior work addressing the limitations of recency-based eviction and the performance degradation caused by constrained memory into two groups.

\vspace{0.5mm}

\noindent \textbf{(a) Eviction-based Policies.} Several works attempt to mitigate the limitations of recency-based eviction in SWA including, AdaKV~\cite{feng2024ada} introduces an adaptive, plug-and-play KV-budget allocation mechanism that integrates with existing cache management policies. 
MSWA~\cite{xu2025mswa} instead varies the attention window size $w$ across heads and layers to improve computational efficiency while maintaining model performance. \textit{Although saliency-based eviction mitigates the limitations of recency-based policies, it often incurs noticeable performance degradation as a consequence of discarding cached entries.}


\vspace{0.5mm}

 \noindent \textbf{(b) Retention-based Policies.} These works address the performance degradation caused by eviction-based policies under constrained memory by incorporating a linear recurrent state. LESS~\cite{dong2024get} introduces a sparse KV-cache strategy where discarded KV pairs are absorbed into a recurrent state via linear attention with an MLP-based $\phi$-kernel. BASED~\cite{arora2024simple} combines sliding-window attention with a degree-2 Taylor approximation of the softmax to realize linear attention while preserving global context. Infini-Transformer~\cite{munkhdalai2024leave} processes inputs in segments while maintaining a recurrent state using ELU-based linear attention, enabling effectively unbounded context under limited memory. EdgeInfinite~\cite{chen2025edgeinfinite} follows a similar segmented design but introduces a learnable gate to regulate the contribution of ELU-based linear attention. \textit{Despite the performance enhancement via constant recurrent state, these policies often requires extensive re-training.} 
 
 \subsection{Parameter-Efficient Hybrid Architecture}

The inefficiency introduced by the complex training procedures in prior hybrid architectures~\cite{arora2024simple, dong2024get} can be mitigated by adopting parameter-efficient fine-tuning (PEFT) strategies. In particular, Low-Rank Adaptation (LoRA)~\cite{hu2022lora} enables efficient adaptation by injecting trainable low-rank matrices into frozen pretrained weights, dramatically reducing memory and training cost. LoLCats~\cite{zhang2025lolcats} employs this LoRA-based approach to distill pretrained softmax attention into learnable $\phi$-kernel linear attention, while Liger~\cite{lan2025liger} applies the same strategy for non-learnable $\phi$-kernel linear attention, establishing a practical path for hybrid architecture adaptation without full retraining. However, despite prior works on adaptive KV-cache budgeting~\cite{feng2024ada, xu2025mswa, cai2024pyramidkv}, we observe that most hybrid architectures still employ a uniform hybridization strategy across layers. \textit{This raises a fundamental question: is uniform layer-wise hybridization optimal, or can non-uniform $\delta$-allocation exploit depth-dependent linearization tolerance?}

\section{Methodology} \label{sec:methods}

In this section, we present \textbf{\textsc{Glide}}, a
\textbf{G}uided \textbf{L}ayerwise Hybrid Attention that strategically integrates
sliding-window softmax attention with linear recurrent aggregation. We first present our motivating observations (Section~\ref{sec: perf}) through an empirical study in which we systematically vary the ratio of softmax-to-linear attention across transformer layers and measure the resulting impact on downstream task accuracy, which reveal the layer-wise heterogeneity in softmax importance and tolerance to linearization. We then present the \textbf{\textsc{Glide}} framework in detail (Section~\ref{sec:glide_method}) as well as the configuration strategies (Section~\ref{sec:layer_alloc}) to exploit the layer-wise heterogeneity.

\subsection{Motivating Observations} \label{sec: perf}


Prior work has shown that the quality degradation inherent to attention linearization can be partially recovered through selective softmax augmentation~\cite{zhang2025lolcats}. However, the optimal allocation of softmax computation remains an open problem: excessive softmax reintroduces the KV cache bottleneck that linearization seeks to eliminate, while insufficient softmax compromises model expressivity. This raises a fundamental question: \textit{What is the minimal softmax budget required to maintain downstream quality under linearization, and is this requirement uniform across layers or depth-dependent?}
\vspace{1mm}

\noindent \textbf{Accuracy Recovery from Linearization.} To understand how softmax attention affects model quality, we analyze the Liger~\cite{lan2025liger} model \textcolor{red}{with Llama-3~8B architecture as backbone} by progressively reintroducing softmax attention from a fully linearized baseline without any fine-tuning, as shown in Fig.~\ref{fig:abs_1}(a). Fig.~\ref{fig:abs_1}(b), ``All Layers'' traces this recovery: zero-shot accuracy starts at 36\% under full linearization and steadily climbs as softmax is uniformly reintroduced, reaching 72\% in the softmax-only configuration. Interestingly, the recovery curve shows diminishing returns, suggesting that only a sparse softmax budget may be needed to restore model quality while retaining most efficiency gains, and to assess the generalizability of this behavior, we also evaluate the Liger setup using a Mistral backbone; results are presented in Fig.~\ref{fig:abs_adx_1} (Appendix~\ref{sec:appd}). This leads to a natural question: if only a fraction of softmax computation is required, can we allocate it selectively rather than spreading it uniformly? We hypothesize that Transformers exhibit non-uniform sensitivity to linearization along the depth axis, with early layers acting as critical representational bottlenecks while deeper layers possess enough redundancy to tolerate linear approximation. To test this, we analyzed our baseline LLM using a pairwise isolated linearization setup, where a pair of layers from a specific region (early, middle, or late) is linearized while all other layers retain full softmax attention. As Fig.~\ref{fig:abs_1}(b) reveals, linearizing early layers $\{1, 2\}$ causes a catastrophic accuracy drop to roughly 36\%, whereas linearizing middle $\{16, 17\}$ and late $\{31, 32\}$ layers produces performance nearly indistinguishable from the full-softmax baseline. This finding, that tolerance to linearization grows with layer depth, forms the foundation of the \textbf{\textsc{Glide}} framework. Rather than applying a uniform policy, \textbf{\textsc{Glide}}  selectively prioritizes softmax attention in early-stage bottlenecks while maximizing linearization in deeper layers, preserving model utility while reducing KV-cache overhead.

\begin{takeawaybox}
\textbf{Observation 1:} Transformer layers exhibit \textbf{depth-dependent sensitivity to attention linearization:} early layers require high softmax intensity to preserve accuracy, while deeper layers tolerate aggressive linearization with minimal degradation.
\end{takeawaybox}

\begin{figure*}[t]
    \centering
    \includegraphics[width=\textwidth]{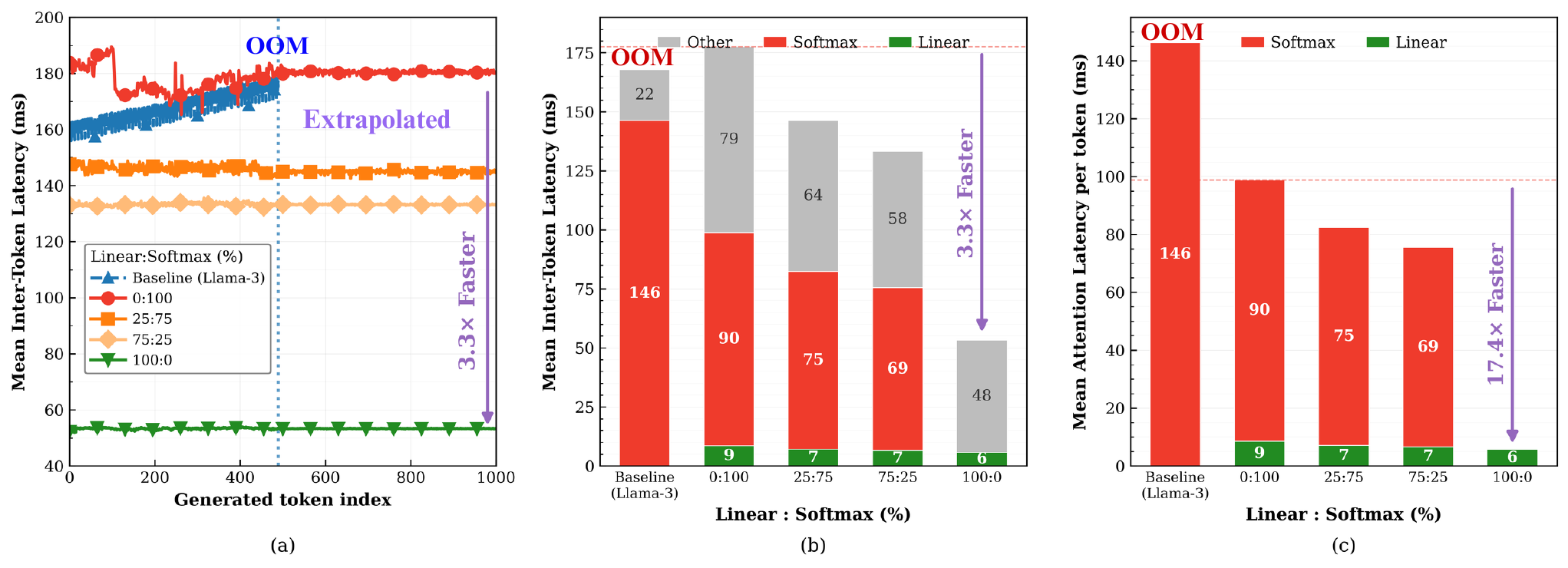}
    \caption{
        \textbf{Attention Latency Analysis Across Linearization Ratios and Baseline.} Profiling Llama-3-8B decoding from 20K-token prefill across uniform linear:softmax ratios with the base window size $w=20\text{K}$ and model dimension $d=4096$. (a) Baseline encounters OOM mid-sequence while GLIDE configurations complete decoding with stable latency, achieving $3.3\times$ speedup at full linearization. (b) Operator breakdown: softmax (red) dominates baseline cost and triggers OOM at 3K tokens; increasing linear ratio eliminates this bottleneck while non-attention overhead (gray) remains constant. (c) Attention-only latency: linear layers contribute ~7ms regardless of sequence length, yielding $17.4\times$ reduction versus softmax.
    }
    \label{fig:motiv_latency_breakdown}
\end{figure*}

\begin{figure}[t!]
    \centering
    \includegraphics[width=0.95\columnwidth]{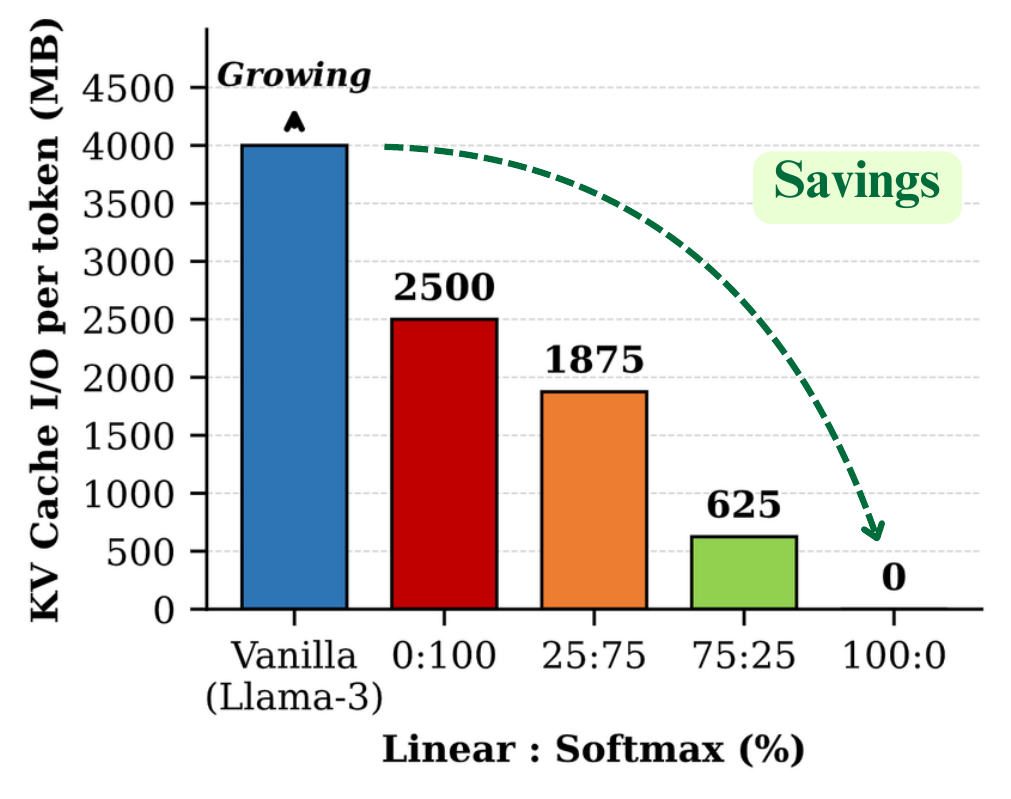}
    \caption{
        \textbf{KV Cache I/O Scales Linearly with Softmax Layer Proportion.} Per-token KV cache memory traffic for Llama-3-8B at $32,000^{th}$ token generation for the window size $w=\text{20k}$ and the model dimension $d=\text{4096}$. Softmax layers retrieve the full KV cache per token; linear layers maintain fixed-size recurrent states with zero cache I/O. Transitioning from pure softmax (0:100) to 75:25 linear:softmax yields a $4\times$ reduction in memory bandwidth.
    }
    \label{fig:mot_cache}
\end{figure}

\vspace{1mm}

\noindent \textbf{Low KV Cache I/O with Reduced FLOPs.} The structural non-uniformity identified in our performance analysis (Section~\ref{sec: perf}) suggests a significant opportunity to optimize LLM inference by reducing both computational latency and memory footprint. To quantify these potential efficiency gains, we evaluate a hybrid attention architecture across varying KV-cache retrieval intensities uniformly distributed across layers. Rather than the growing memory complexity inherent in standard Transformers, this selective retrieval strategy leverages the constant-size recurrent states of linear layers to mitigate the I/O bottleneck. As illustrated in Fig.~\ref{fig:mot_cache}, the per-token KV-cache I/O scales linearly with the proportion of softmax layers. For the Llama-3 8B baseline (0:100 linear:softmax ratio), memory movement peaks at approximately 4000 MB per token at 32,000 tokens of context. Transitioning to a 25:75 configuration reduces this to 1875 MB, while a 75:25 ratio yields just 625 MB, achieving a 4$\times$ reduction in memory bandwidth. In the extreme 100:0 linear configuration, the per-token I/O overhead for historical context is eliminated entirely. These results highlight that even modest increases in linearization yield substantial memory savings, motivating a depth-aware allocation strategy that balances accuracy with I/O efficiency.

\begin{takeawaybox}
\textbf{Observation 2:} Hybrid attention achieves \textbf{ memory bandwidth reduction}. Even modest linearization (25:75 linear:softmax) yields substantial savings, motivating depth-aware layer-wise allocation.
\end{takeawaybox}


\noindent \textbf{Improved Latency Across Configurations.} This reduction in I/O translates directly to improved temporal performance. Fig.~\ref{fig:motiv_latency_breakdown} illustrates the impact on inter-token latency, driven by the compression of floating-point operations within the attention mechanism. The mean inter-token latency drops from approximately 180ms in the baseline to roughly 55ms in the linearized setup, representing a 3.3$\times$ speedup. A deeper breakdown of the attention component reveals that while the linear attention latency remains minimal and constant (7--9ms), the softmax overhead is the primary driver of execution time. Notably, the \textbf{\textsc{Glide}} framework's ability to selectively linearize high-redundancy layers allows for a significant breach of the "memory wall" while preserving the representational fidelity of the early-stage bottleneck layers. By leveraging architectural insights into attention mechanisms and the efficiency characteristics revealed through observed geometric properties, \textbf{\textsc{Glide}} constitutes an adaptive framework tailored specifically for hybrid attention architectures. This formulation enables extended context processing while substantially reducing KV cache I/O traffic, consequently minimizing per-token FLOP requirements and yielding lower inter-token latency during the decode phase.

\begin{takeawaybox}
\textbf{Observation 3:}  Partial linearized layers can achieve a significant \textbf{reduction in KV-cache I/O} and \textbf{improves overall speedup}. This strategic retrieval yields a notable localized acceleration in attention computation while preserving critical early-layer bottlenecks.
\end{takeawaybox}

\subsection{GLIDE Attention} \label{sec:glide_method}


\textbf{\textsc{Glide}} is designed around two key observations: (1)~transformer layers exhibit non-uniform sensitivity to linearization, and (2)~significant efficiency gains are achievable across various retrieval rates, \textbf{\textsc{Glide}} functions as an adaptive module that augments existing hybrid attention mechanisms, facilitating extended context processing with improved computational efficiency relative to baseline architectures. \textbf{\textsc{Glide}} mitigates information loss arising from selective retrieval by applying linear attention to the $\delta$ region within the sliding window $w$. This mechanism enables controlled linearization governed by the choice of $\delta$, concentrating local attention while reducing computational overhead with minimal impact on downstream performance. The selection of $\delta$ is guided by the observed behavior of the attention mechanism, permitting increased linearization in deeper layers where sensitivity to approximation is lower. Furthermore, tokens evicted from the sliding window are seamlessly incorporated into the linear recurrent memory, preserving information without increasing memory footprint as the sequence length scales. \textbf{This mechanism underpins the ability of \textbf{\textsc{Glide}} to process extended contexts efficiently with bounded memory requirements.} The proposed framework is a modular and adaptive framework designed for hybrid attention architectures that combine sliding window attention (SWA) with linear attention. The framework is architecture-agnostic, supporting a variety of linear attention variants including Taylor-based kernels \cite{arora2024simple},  \cite{zhang2402hedgehog}, and GLA-style gating mechanisms \cite{lan2025liger}. Crucially, \textbf{\textsc{Glide}} operates with any non-learnable $\phi$-kernel, enabling parameter-efficient deployment without requiring additional fine-tuning or learned projections. This design choice ensures low-latency inference while maintaining compatibility with pretrained transformer weights. \textit{Given a query $q_i$ at the instance $i$, the output of \textbf{\textsc{Glide}} attention can be defined as,}

\begin{equation} \label{eq:7}
O^{\text{Glide}}_i = O^{\text{linear}}_{i-w-\delta} \;\oplus\; O^{\text{SWA}}_{i-w-\delta+1},
\end{equation}
where $w$ denotes the constant KV cache size of Sliding Window Attention (SWA), and $\delta$ is the parameter controlling selective KV cache retrieval from this window. The parameter $\delta$ effectively determines the degree of linearization applied at each layer: a larger $\delta$ implies greater reliance on linear attention, while $\delta = w$ corresponds to full linearization, completely bypassing KV cache retrieval. In practice, setting $\delta < w$ reduces the effective KV I/O from $w$ to $w' = w - \delta$, enabling fine-grained control over the trade-off between memory bandwidth and attention fidelity. Further breaking down \eqref{eq:7}, the concentrated local attention computed by Sliding Window Attention (SWA) can be given as, 

\begin{equation}
     O^{\text{SWA}}_{i-w-\delta+1} = \frac{\sum_{j'=i-w-\delta+1}^{i}\exp(q_i k_{j'}^T) \cdot v_{j'}}{\sum_{j'=i-w-\delta+1}^{i} \exp(q_i k_{j'}^T)},
\end{equation}

\noindent Tokens evicted from the sliding window KV cache, together with the $\delta$ most recent tokens within the window, are handled via linear attention rather than full softmax attention. This design enables efficient global context modeling through a constant-size recurrent state, eliminating the need to store or retrieve explicit key-value pairs for tokens outside the active window. The linear attention component can be formally defined as,

\begin{equation}
  O_{i-w-\delta}^{\text{Linear}} = \frac{\sum_{j'=1}^{i-w-\delta}\phi(q_i)\cdot\phi(k_{j'})^T \cdot v_{j'}}{\sum_{j'=1}^{i-w-\delta}\phi(q_i)\cdot\phi(k_{j'})^T},
\end{equation}

\noindent where $\phi(\cdot)$ denotes a non-learnable separable kernel function that enables parameter-efficient inference without requiring additional learned projections. In our implementation, we adopt $\phi(\cdot) = \text{softmax}(\cdot)$ as the default kernel choice, following recent findings that softmax-based feature maps preserve the spectral properties of the original attention mechanism more effectively than polynomial or RBF alternatives. The \textbf{\textsc{Glide}} architecture preserves the full representational capacity of softmax attention by combining two complementary components: adaptive local attention over the sliding window, which captures fine-grained token interactions within a bounded context, and constant-memory global attention via linear recurrence, which maintains a compressed summary of the entire prefix history. The combination operator $\oplus$ merges these two attention outputs, weighted by their respective normalizing denominators, ensuring proper probability mass allocation across the full context and preventing either component from dominating the final output. As the hybridization is generic, \textsc{Glide} supports various  formulations of the hybrid attention works, including BASED, LoLCats~\cite{arora2024simple, zhang2025lolcats} and Liger~\cite{lan2025liger} (see  Appendix~\ref{sec:swa}). This formulation guarantees two important boundary conditions: when $w \to \infty$, the sliding window spans the entire sequence and \textbf{\textsc{Glide}} attention reduces to standard causal softmax attention, and when $w = 0$, the mechanism collapses to pure linear attention with recurrent state updates, ensuring that \textbf{\textsc{Glide}} generalizes both attention paradigms within a unified framework. We formally define the [Linear:Softmax] ratio in our previous illustrations as a \textbf{\textsc{Glide}} $\delta$-allocation as below:

\noindent\begin{itemize}[leftmargin=*, itemsep=0pt]
    \item $\boldsymbol{\delta} = \{0, 0,\ldots,0\}$, i.e., $\delta_\ell=0$ for all $l$: This corresponds to the baseline \textbf{Hybrid} configuration. Specifically, within the sliding window of size $w$, all layers allocate $0\%$ to linear attention and $100\%$ to softmax attention. The KV cache stores only softmax states.
    
    \item $\boldsymbol{\delta} = \{w, w,\ldots,w\}$, i.e., $\delta_l=w$ for all $l$: This corresponds to the baseline \textbf{Linear} configuration. Within the sliding window $w$, all layers allocate $100\%$ to linear attention and $0\%$ to softmax attention. The KV cache stores only linear compressed states.

    \item $\boldsymbol{\delta} = \{0,\ldots,0~|~\alpha \cdot w,\ldots,\alpha \cdot w~|~w,\ldots,w\}$: This corresponds to a  \textbf{\textsc{Glide}} configuration discussed in the next section. Here, different layer groups are assigned varying ratios of linear and softmax attention within the window $w$. The parameter $\alpha \in [0,1]$ is defined as cache sparsity, where, $\delta_l = \alpha \cdot w$ means the layer uses $\alpha\cdot w$ (gliding) tokens from cache for linear attention and remaining $(1-\alpha)w$ tokens for softmax attention incurring KV cache I/O overhead for the latter.
    For instance, $\boldsymbol{\delta} = \{0,\ldots,0~|~w/2,\ldots,w/2~|~w,\ldots,w\}$ represents a three-block configuration where early layers use pure softmax ($\alpha=0$, no linear attention, no sparsity), middle layers use equal mix ($\alpha=0.5$, yielding $50\%$ linear and $50\%$ softmax), and last layers use pure linear ($\alpha=1$, no softmax, aggressive sparsity). This configuration is denoted as $(0, w/2, w)$. 
    
    
\end{itemize}

\begin{algorithm}[ht]
\caption{Blockwise $\delta$ Allocation}
\label{alg:block_alloc}
\small

\KwIn{Model $M$, window $w$}
\KwOut{Optimized model $M^*$}

$(B_1,B_2,B_3) \leftarrow \textsc{Split}(M, 3)$\;
\textsc{Assign}($B_1,0$)\;
\textsc{Assign}($B_3,w$)\;
$\mathcal{S} \leftarrow \{0,\tfrac{w}{2},\tfrac{15w}{16},w\}$\;

\ForEach{$\delta \in \mathcal{S}$}{
    \textsc{Assign}($B_2,\delta$)\;
}

$\delta^* \leftarrow \arg\max_{\delta \in \mathcal{S}} \textsc{TaskAccEval}(M_\delta)$\;
\textsc{Assign}($B_2,\delta^*$)\;
\Return $M^*$\;
\end{algorithm}

\vspace{1mm}

\noindent \textbf{Sub-quadratic Efficiency.}
Under autoregressive decoding, the resulting attention operator $O^{\mathrm{Glide}}_i$ attains a per-token time and space complexity of $\mathcal{O}((\delta_\ell) d + d^2)$, where $d$ is the head dimension and $\delta_\ell$ denotes the layer-specific linearization parameter at layer $\ell$. Since $\delta_\ell$ controls the extent of linearization within the window, larger values reduce the effective quadratic interaction region from $w$ to $(w - \delta_\ell)$, progressively replacing exact pairwise attention with linearized aggregation. In the extreme case $\delta_\ell = w$, layer $\ell$ becomes fully linear with respect to the window, whereas $\delta_\ell = 0$ recovers standard sliding-window attention. Consequently, for an $L$-layer model with layer-wise allocations $\boldsymbol{\delta} = \{\delta_1, \delta_2, \ldots, \delta_L\}$, the total per-token decoding complexity becomes $\mathcal{O}\left(\sum_{\ell=1}^{L} (w - \delta_\ell) d + L d^2\right)$, which simplifies to $\mathcal{O}\left((Lw - \sum_{\ell=1}^{L} \delta_\ell) d + L d^2\right)$. This formulation reveals that \textbf{\textsc{Glide}} provides a continuous and controllable interpolation between quadratic and linear regimes, enabling explicit accuracy--efficiency trade-offs with sub-quadratic decoding complexity. Furthermore, leveraging non-uniform allocation across layers, where deeper layers typically admit larger $\delta_\ell$ values due to reduced sensitivity to linearization, \textit{\textbf{\textsc{Glide}} achieves near-minimal complexity in later layers while preserving representational fidelity in early layers where precise attention is critical.}

\begin{figure}[ht]
    \centering
    \includegraphics[width=\columnwidth]{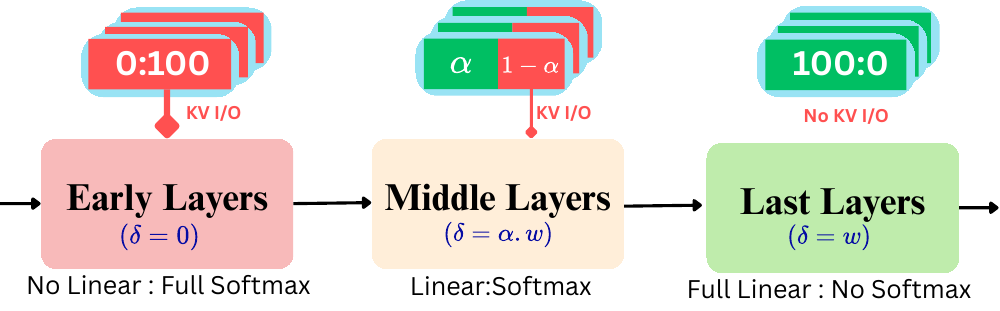}
    \caption{
       \textbf{Proposed $\delta$-Allocation Strategy for \textbf{\textsc{Glide}}.} Block-wise allocation partitions the model into blocks of early, middle, and last layers based on our motivating Observation 1 (in Section~\ref{sec: perf}). Early layers retain full softmax attention ($\delta = 0$) where representational fidelity is critical, last layers use complete linearization ($\delta = w$) for maximum efficiency, and middle layers exhibit glide attention with $\delta = \alpha \cdot w$ where $\alpha \in [0,1]$, enabling fine-grained control over the efficiency-accuracy trade-off.
    }
    \label{fig:alloc}
\end{figure}

\begin{table*}[t]
\caption{\textbf{Fine-Tuned Accuracy Across Reasoning and Knowledge Benchmarks.} We evaluate Fine-tuned GLIDE across standard benchmarks: PiQA, ARC-Easy, ARC-Challenge, HellaSwag, and WinoGrande in zero-shot settings, with MMLU evaluated using 5-shot prompting. Each row corresponds to a block-wise $\delta$-configuration ranging from full softmax retention $(0,0,0)$ to complete linearization $(w,w,w)$ with the base window size $w=1204$ with model dimension $d=4096$. Avg. denotes the mean across all six tasks. Results show that moderate linearization retains competitive performance.}
\label{tab:fine_tuned}


\centering
\resizebox{\textwidth}{!}{
\renewcommand{\arraystretch}{1.2}
\begin{tabular}{|l|l|cccccc|>{\columncolor{shade}}c|>{\columncolor{shade}}c|}
\hline
\multirow{2}{*}{\textbf{Model}} & 
\multirow{2}{*}{\textbf{Configuration}} & 
\multirow{2}{*}{\textbf{PiQA} ($\uparrow$)} & 
\multirow{2}{*}{\textbf{ARC-E} ($\uparrow$)} & 
\multirow{2}{*}{\textbf{ARC-C} ($\uparrow$)} & 
\multirow{2}{*}{\textbf{H.Swag} ($\uparrow$)} & 
\multirow{2}{*}{\textbf{WinoG.} ($\uparrow$)} & 
\multicolumn{1}{c}{\textbf{MMLU} ($\uparrow$)} & 
\multicolumn{1}{|>{\columncolor{shade}}c|}{\textbf{Avg.} ($\uparrow$)} & 
\multicolumn{1}{|>{\columncolor{shade}}c|}{\textbf{KV I/O} ($\downarrow$)} \\

& & & & & & & \multicolumn{1}{c}{\footnotesize\textbf{(5-shot)}} & 
\multicolumn{1}{|>{\columncolor{shade}}c|}{} & 
\multicolumn{1}{|>{\columncolor{shade}}c|}{\footnotesize\textbf{(MB/tok)}} \\ 
\hline

\multirow{6}{*}{Llama-3-8B} 
& Baseline (Vanilla SoftMax) & $78.84$ & $80.98$ & $53.92$ & $79.19$ & $73.32$ & $66.68$ & $72.16$ & $4000$ \\
& \textbf{\textsc{Glide}}~$(0,0,0) \equiv \text{Baseline (Hybrid)}$ & $80.20$ & $79.88$ & $51.71$ & $77.41$ & $71.11$ & $54.39$ & $69.12$ & $128$ \\
& \textbf{\textsc{Glide}}~$(0,0,w)$ & $78.73$ & $78.58$ & $49.91$ & $76.03$ & $70.09$ & $53.68$ & $67.84$ & $88$ \\
& \textbf{\textsc{Glide}}~$(0,w/2,w)$ & $78.89$ & $78.11$ & $50.34$ & $75.93$ & $70.48$ & $52.62$ & $67.73$ & $64$ \\
& \textbf{\textsc{Glide}}~$(0,15w/16,w)$ & $78.24$ & $77.65$ & $48.55$ & $74.58$ & $70.48$ & $50.93$ & $66.74$ & $43$ \\
& \textbf{\textsc{Glide}}~$(w,w,w) \equiv \text{Baseline (Linear)}$ & $53.16$ & $27.57$ & $24.23$ & $25.22$ & $50.51$ & $23.09$ & $33.96$ & $0$ \\
\hline
\multirow{6}{*}{Mistral-7B} 
& Baseline (Vanilla SoftMax) & $80.79$ & $80.22$ & $54.01$ & $81.17$ & $75.30$ & $63.78$ & $72.55$ & $4000$ \\
& \textbf{\textsc{Glide}}~$(0,0,0) \equiv \text{Baseline (Hybrid)}$ & $81.34$ & $81.73$ & $55.12$ & $80.72$ & $72.61$ & $56.32$ & $71.31$ & $128$ \\
& \textbf{\textsc{Glide}}~$(0,0,w)$ & $80.63$ & $79.76$ & $51.45$ & $78.00$ & $73.24$ & $55.69$ & $69.80$ & $88$ \\
& \textbf{\textsc{Glide}}~$(0,w/2,w)$ & $80.30$ & $80.05$ & $51.19$ & $77.98$ & $73.16$ & $53.93$ & $69.44$ & $64$ \\
& \textbf{\textsc{Glide}}~$(0,15w/16,w)$ & $80.41$ & $79.97$ & $51.28$ & $76.08$ & $73.16$ & $48.48$ & $68.23$ & $43$ \\
& \textbf{\textsc{Glide}}~$(w,w,w) \equiv \text{Baseline (Linear)}$ &  $53.26$ & $28.49$  & $24.15$ & $25.92$ & $51.14$ & $23.09$ & $34.34$ & $0$ \\
\hline
\end{tabular}
}
\end{table*}

\subsection{Sensitivity-Guided Layerwise Allocation} \label{sec:layer_alloc}

\textbf{\textsc{Glide}} attention supports fine-grained layer-wise allocation of $\delta_\ell$, enabling per-layer control over the linearization--fidelity trade-off to maximize throughput while preserving downstream task performance. However, fully heterogeneous layer-wise configurations can incur runtime overhead due to frequent kernel switching and reduced operator fusion opportunities. To address this, we introduce block-wise $\delta$ allocation, where contiguous groups of $B$ layers share a common $\delta$ value, i.e., $\delta_\ell = \delta_b$ for $\ell \in \mathcal{B}_b$. This coarser granularity reduces configuration complexity while still capturing the depth-dependent sensitivity patterns observed in Section~\ref{sec: perf} namely, that early layers require smaller $\delta$ values for representational fidelity, while deeper layers tolerate aggressive linearization. Block-wise allocation also enables kernel-aware pipelining and configuration reuse across layers, improving wall-clock efficiency with minimal impact on model accuracy. 

We propose a simplified block-wise allocation strategy described in Algorithm~\ref{alg:block_alloc} to reduce the search space. Here we partition the model into blocks of early, middle, and last layers as illustrated in Fig.~\ref{fig:alloc} and \textsc{SPLIT} function in Algorithm~\ref{alg:block_alloc}.
The early block is assigned $\delta_{b=1} = 0$ (full sliding window softmax attention) and the last block is assigned $\delta_{b=3} = w$ (full linear attention). The middle block allows us to allocate $\delta_{b=2}=\alpha \cdot w$, where, $\alpha \in [0, 1]$, is cache sparsity, and to explore the performance–efficiency tradeoffs across various \textsc{\textbf{Glide}} configurations, we restrict $\alpha$ such that $\delta \in \{0,\, w/2,\, 15w/16,\, w\}$ within the search space described in Algorithm~\ref{alg:block_alloc}. 

\noindent
\textbf{Speedup.} \textit{For three equally-sized blocks, each comprising $L/3$ layers, we achieve a theoretical computational speedup of} 
$$S=\frac{Lwd + Ld^2}{(Lwd/3 + Lw(1 - \alpha)d/3 + Ld^2)} 
= 1 + \frac{1+\alpha}{(2 +3d/w)- \alpha}$$ 
\textit{for per-token~decoding with respect to baseline hybrid model, which is a specific configuration,} \textbf{\textsc{Glide}}~(0,0,0) with $\alpha=0$. The theoretical speedup $S$ is highly context-dependent, scaling up to a $3\times$ maximum in the long-context regime ($w \gg d$) where softmax attention acts as the primary computational bottleneck and is highly responsive to cache sparsity $\alpha$. 

\begin{figure*}[t]
    \centering
    \subfloat[Llama-3-8B\label{fig:acc_kv_llama}]{%
        \includegraphics[width=0.45\textwidth]{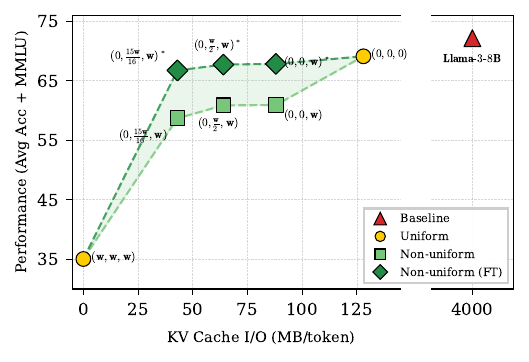}}
    \hspace{1.5em}
    \subfloat[Mistral-7B\label{fig:acc_kv_mistral}]{%
        \includegraphics[width=0.45\textwidth]{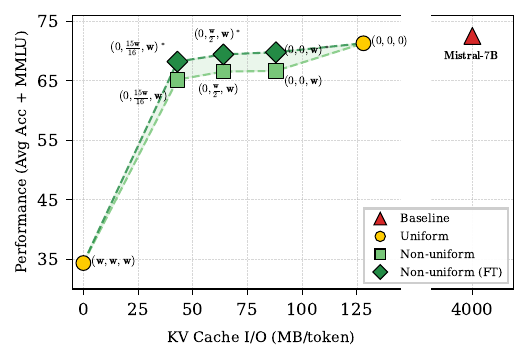}}
    \caption{\textbf{Performance vs. KV Cache I/O Trade-off.} Average task accuracy versus KV cache I/O per token for (a) Llama-3-8B and (b) Mistral-7B. Markers indicate: baseline ({\large\textcolor{red}{$\blacktriangle$}}), uniform ({\large\textcolor{yellow}{$\bullet$}}), non-uniform ({\large\textcolor{green!60}{$\blacksquare$}}), and non-uniform fine-tuned ({\large\textcolor{green!40!black}{$\blacklozenge$}}). Fine-tuning recovers 6--8 accuracy points, achieving 96\% baseline performance at 45× lower memory bandwidth while non-uniform allocations consistently outperform uniform alternatives.}
    \label{fig:acc_vs_kv_io}
\end{figure*}

\section{Experiments} \label{sec:glide_exp}

This section validates ability of \textbf{\textsc{Glide}} to exploit layer-wise linearization for reducing KV cache I/O overhead and accelerating inference while preserving downstream task accuracy. We systematically evaluate computational efficiency through two primary metrics: (1) KV cache I/O overhead, quantified as the required total KV cache movement, and (2) end-to-end inference latency, measured across sequence lengths. Task performance is assessed on six standard reasoning and knowledge benchmarks from the language model evaluation harness: PiQA (physical commonsense), ARC-Easy and ARC-Challenge (science reasoning), HellaSwag (commonsense inference), WinoGrande (coreference resolution), and MMLU (multitask language understanding). To address potential accuracy degradation under aggressive linearization, we apply LoRA-based parameter-efficient fine-tuning (PEFT) to \textbf{\textsc{Glide}} configurations, enabling recovery of performance while maintaining the reduced KV cache footprint of linearized layers. Our experimental analysis addresses the following research question: \textit{Can \textbf{\textsc{Glide}}'s per-layer $\delta$-allocation mechanism reduce KV cache I/O and computational cost while preserving downstream task accuracy through strategic layer-wise linearization?}




\subsection{Experimental Setup}

\textbf{Implementation Configurations.} Our implementation builds upon Liger and Lolcats~\cite{lan2025liger, zhang2025lolcats} and leverages PyTorch's FlexAttention with Flash Attention Backend~\cite{dong2025flexattention} to realize an adaptive sliding-window attention mechanism. We conducted our experiments on a single NVIDIA Grace Hopper (GH200 Superchip: Grace CPU + Hopper GPU) 120GB, and we fine-tuned the model using LoRA with a rank of 8 and a scaling coefficient of 8. Training is performed for two epochs on 100K cleaned Alpaca instruction samples ($\approx$ 0.02B tokens) to support gate-recurrent linearization. All sequences are set to a maximum length of 1024 tokens in the LM eval benchmark framework. Optimization is carried out with a micro-batch size of 1, and an effective batch size of 8 is realized via gradient accumulation, consistent with the LoLCATs training setup \cite{zhang2025lolcats} with base window size $w=1024$. 

\noindent \textbf{Models \& Datasets.} To systematically quantify the layer-wise sensitivity of \textbf{\textsc{Glide}}, we perform a controlled paired-replacement study across varying $\delta$ configurations, isolating the effect of selective linearization at each transformer layer.  Fine-tuning is conducted for two epochs on 50{,}000 cleaned Alpaca~\cite{alpaca} instruction-following samples to ensure stable adaptation under limited supervision. To demonstrate generality, we integrate \textbf{\textsc{Glide}} into strong, widely adopted backbone models such as LLaMA~3~8B~\cite{grattafiori2024llama3herdmodels} and Mistral~7B~\cite{jiang2023mistral7b} with baseline model dimension $d=4096$, which serve as competitive baselines for comparative evaluation. Following the sensitivity hierarchy illustrated in Fig.~\ref{fig:layer_sensitivity}, 
we partition both models ($L = 32$, 0-indexed) into early 
(layers $1$--$11$, $\delta = 0$), middle (layers $12$--$21$, 
$\delta = \alpha \cdot w$), and late (layers $22$--$32$, $\delta = w$) 
blocks, assigning conservative softmax budgets 
where sensitivity is highest and aggressive linearization where redundancy 
is established.

\subsection{Results and Discussion}
To validate the efficiency-performance trade-off enabled by \textbf{\textsc{Glide}}, we evaluate the framework on six standard reasoning benchmarks: PiQA, ARC-Easy, ARC-Challenge, HellaSwag, WinoGrande, and MMLU. We compare \textbf{\textsc{Glide}} configurations against three baselines: vanilla softmax attention, hybrid attention, and pure linear attention. Efficiency gains are measured through two metrics: end-to-end latency and KV cache I/O volume across varying sequence lengths. These results demonstrate the Pareto-optimal configurations enabled by \textbf{\textsc{Glide}} framework. 

\vspace{1mm}

\noindent \textbf{Recovering Performance through PEFT.} 
When \textbf{\textsc{Glide}} configurations are deployed as drop-in replacements for standard attention with various non-uniform $\delta$-allocations, we observe a degradation of $5\%$ to $10\%$ in downstream task performance relative to the baseline hybrid model. This performance gap is visualized in Fig.~\ref{fig:acc_vs_kv_io}(a) and (b) as the vertical distance between the $(0, 0, 0)$ configuration ({\large\textcolor{yellow}{$\bullet$}}, baseline hybrid with full softmax in all blocks) and the $(0, 0, w)$ configuration ({\large\textcolor{green!60}{$\blacksquare$}}, aggressive linearization in later blocks). While the vanilla softmax baseline ({\large\textcolor{red}{$\blacktriangle$}}) achieves the highest accuracy of approximately $72\%$ on both models, it incurs a prohibitive KV cache I/O cost of 4000 MB/token. In contrast, the $(0, 0, 0)$ hybrid baseline achieves $69$--$71\%$ accuracy at only 128 MB/token, representing a $31\times$ reduction in memory bandwidth with minimal performance loss. To recover the performance degradation observed in more aggressive linearization schemes while maintaining efficiency gains, we apply LoRA-based parameter-efficient fine-tuning (PEFT) to the \textbf{\textsc{Glide}} configurations. Fine-tuned models are indicated by dark green diamond markers ({\large\textcolor{green!40!black}{$\blacklozenge$}}) in Fig.~\ref{fig:acc_vs_kv_io}. The results demonstrate that fine-tuning substantially closes the accuracy gap: the fine-tuned $(0, 0, w)$ configuration recovers $6$ to $8$ accuracy points, reaching approximately $68$--$70\%$ accuracy (96\% of baseline performance) while operating at only 88 MB/token—a $45\times$ reduction compared to vanilla softmax. Similarly, the $(0, 15w/16, w)$ configuration achieves $66$--$68\%$ accuracy at just 43 MB/token, nearly $93\times$ more efficient than the vanilla baseline. \textit{This pattern holds across all non-uniform $\delta$-allocations, with fine-tuned variants ({\large\textcolor{green!40!black}{$\blacklozenge$}}) consistently outperforming their zero-shot counterparts ({\large\textcolor{green!60}{$\blacksquare$}}) at equivalent KV cache I/O levels, demonstrating \textbf{\textsc{Glide}}'s ability to unlock a rich Pareto frontier of efficiency-performance trade-offs.}

\vspace{1mm}

\noindent \textbf{Performance Trade-offs Across \textbf{\textsc{Glide}} Configurations.} 
Table~\ref{tab:fine_tuned} presents fine-tuned accuracy results for Llama-3-8B and Mistral-7B across six reasoning and knowledge benchmarks. The vanilla softmax baseline achieves 72.16\% average accuracy on Llama-3-8B and 72.55\% on Mistral-7B, but requires 4000 MB/token KV cache I/O. The hybrid baseline configuration $(0, 0, 0) \equiv \text{baseline (Hybrid)}$ achieves 69.12\% and 71.31\% respectively at only 128 MB/token—a $31\times$ reduction in memory bandwidth while retaining 96--98\% of baseline performance. Progressive linearization reveals distinct per-layer sensitivity patterns. For Llama-3-8B, the $(0, 0, w)$ configuration achieves 67.84\% accuracy at 88 MB/token. Extending linearization to intermediate layers with $(0, w/2, w)$ maintains nearly identical performance at 67.73\%, indicating that mid-layer linearization incurs negligible additional cost once the final block is linearized. The aggressive $(0, 15w/16, w)$ configuration yields 66.74\% at just 43 MB/token—a $93\times$ reduction compared to vanilla softmax while preserving 92\% of baseline accuracy. However, full linearization $(w, w, w) \equiv \text{baseline~(Linear)}$ degrades catastrophically to 33.96\%, with severe failures on reasoning tasks such as HellaSwag (25.22\%) and ARC-C (24.23\%). Mistral-7B exhibits consistent trends with stronger absolute performance. Progressive linearization through $(0, 0, w)$, $(0, w/2, w)$, and $(0, 15w/16, w)$ yields 69.80\%, 69.44\%, and 68.23\% respectively, demonstrating graceful degradation of approximately 1\% per linearization step. Full linearization again proves detrimental at 34.34\%, confirming that complete softmax elimination severely impacts model quality across architectures. \textit{These results demonstrate that moderate linearization configurations achieve $45$--$93\times$ efficiency gains while preserving 90--96\% of baseline accuracy after fine-tuning.}

\vspace{1mm}

\begin{table}[t!]
\centering
\begin{threeparttable}
\caption{\textbf{Block-wise $\delta$-Allocation Strategies.} End-to-end accumulated latency over sequence length (S) under \textbf{\textsc{Glide}}'s allocation schemes for Llama-3-8B and Mistral-7B with base window size $w=1024$ and model dimension $d=4096$. Configurations exploit layer-wise linearization tolerance, concentrating softmax in early layers. As $\delta$ increases, memory bandwidth decreases and inference accelerates.} 
\label{tab:allocation_comparison}
\scriptsize
\setlength{\tabcolsep}{2.8pt}
\begin{tabular}{|l|cccc|}
\hline
\multicolumn{5}{|c|}{\textbf{Llama-3-8B}} \\
\hline
\textbf{Configuration} & \multicolumn{4}{c|}{\textbf{E2E Latency (s) over Seq. Length}} \\
\cline{2-5}
  & 4K & 8K & 16K & 32K \\
\hline
    Baseline (Vanilla Softmax) & $264.63$ & $765.92$ & --(OOM)\tnote{1} & --(OOM)\tnote{1} \\
    \textbf{\textsc{Glide}}~$(0, 0, 0) \equiv \text{Baseline}~(\text{Hybrid})$ & $453.68$ & $919.66$ & $1850.92$ & $3705.72$ \\
    \textbf{\textsc{Glide}}~$(0, 0, w)$ & $227.71$ & $453.08$ & $973.58$ & $2638.35$ \\
    \textbf{\textsc{Glide}}~$(0, w/2, w)$ & $223.58$ & $446.23$ & $966.50$ & $2637.42$ \\
    \textbf{\textsc{Glide}}~$(0, 15w/16, w)$ & $225.47$ & $449.67$ & $972.20$ & $2637.54$ \\
    \textbf{\textsc{Glide}}~$(w, w, w) \equiv \text{Baseline}~(\text{Linear})$ & $205.12$ & $409.11$ & $818.59$ & $1640.75$ \\
\hline
\multicolumn{5}{|c|}{\textbf{Mistral-7B}} \\
\hline
\textbf{Configuration} & \multicolumn{4}{c|}{\textbf{E2E Latency (s) over Seq. Length}} \\
\cline{2-5}
  & 4K & 8K & 16K & 32K \\
\hline
    Baseline (Vanilla Softmax) & $305.26$ & $1123.57$ & --(OOM)\tnote{1} & --(OOM)\tnote{1} \\
    \textbf{\textsc{Glide}}~$(0, 0, 0) \equiv \text{Baseline}~(\text{Hybrid})$ & $707.25$ & $1396.80$ & $2789.58$ & $5589.72$ \\
    \textbf{\textsc{Glide}}~$(0, 0, w)$ & $724.06$ & $1448.75$ & $2998.59$ & $5856.51$ \\
    \textbf{\textsc{Glide}}~$(0, w/2, w)$ & $723.66$ & $1476.85$ & $2854.07$ & $5770.93$ \\
    \textbf{\textsc{Glide}}~$(0, 15w/16, w)$ & $679.15$ & $1354.91$ & $2711.12$ & $5444.56$ \\
    \textbf{\textsc{Glide}}~$(w, w, w) \equiv \text{Baseline}~(\text{Linear})$ & $710.38$ & $1415.44$ & $2824.71$ & $5652.72$ \\
\hline
\end{tabular}
\begin{tablenotes}
    \item[1] \scriptsize --(OOM) denotes Out-of-Memory; longer sequence lengths are not supported.
\end{tablenotes}
\end{threeparttable}
\end{table}

\noindent \textbf{End-To-End Accumulated Latency Comparison.}
Table~\ref{tab:allocation_comparison} presents end-to-end latency measurements for Llama-3-8B and Mistral-7B across sequence lengths from 4K to 32K tokens. Llama backbone, the vanilla softmax baseline achieves $264.63$s at 4K but encounters out-of-memory failures beyond 8K tokens. The $(0, 0, 0) \equiv \text{baseline~(hybrid)}$ eliminates these constraints, scaling to $3705.72$s at 32K tokens and enabling long-context inference.Progressive non-uniform linearization yields substantial efficiency gains. The $(0, 0, w)$ configuration achieves $227.71$s at 4K (a 2.0× speedup over the hybrid baseline) and scales to $2638.35$s at 32K. Notably, configurations $(0, 0, w)$, $(0, w/2, w)$, and $(0, 15w/16, w)$ exhibit nearly identical latencies at 32K tokens, completing within $2637$s to $2638$s despite differing $\delta_2$ allocations. This iso-latency behavior arises because memory bandwidth, not computation, dominates the bottleneck. The key differentiator becomes KV cache I/O: these configurations require 88, 64, and $43$ MB/token respectively, representing $2\times$ to $3\times$ memory bandwidth reductions. The fully linearized $(w, w, w) \equiv \text{baseline~(Linear)}$ achieves the lowest latency ($205.12$s at 4K, $1640.75$s at 32K) but suffers catastrophic accuracy degradation to $33.96\%$. Additionally, the Mistral-7B backbone exhibits similar behavior under \textbf{\textsc{Glide}}, where accumulated end-to-end latency shows only minimal improvement across $\delta$-allocation configurations. We attribute this to hardware-level overheads inherent to hybrid attention execution, including warp divergence, irregular data access patterns, and kernel launch overhead. \textit{\textbf{\textsc{Glide}} configurations therefore represent the optimal balance, achieving $1.4\times$ to $2\times$ latency improvements and 2 to 3× memory bandwidth reductions while preserving $92\%$ to $96\%$ of baseline accuracy.}

\vspace{1mm}

\begin{figure}[ht]
    \centering
    \includegraphics[width=\columnwidth]{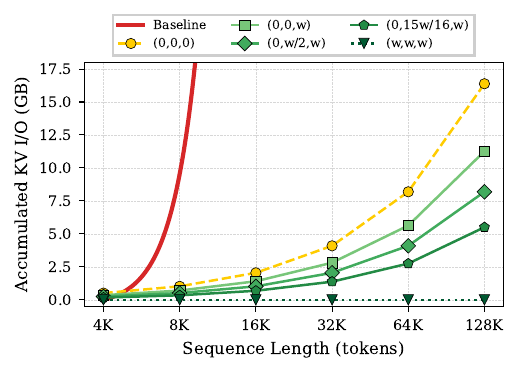}
    \caption{
        \textbf{ Accumulated KV I/O vs. Sequence Length.} \textbf{\textsc{Glide}} configurations achieve sub-linear KV cache growth compared to baseline's linear scaling. Progressive linearization from $(0,0,0)$ to $(0,15w/16,w)$ reduces memory I/O while maintaining accuracy, with $(0,15w/16,w)$ achieving $5.2$ GB at $128$K tokens versus baseline's $16.3$ GB.
    }
    \label{fig:seq_kv_alloc}
\end{figure}

\vspace{1mm}

\noindent \textbf{KV I/O Efficient Reasoning Scalability.} A key challenge in scaling reasoning models to long contexts is the linear growth of KV cache I/O, which shifts inference from being compute-bound to memory-bandwidth-bound. As shown in Fig.~\ref{fig:seq_kv_alloc}, baseline models accumulate over 17.5 GB of KV I/O at 128k tokens, creating a throughput bottleneck where each generated token must retrieve an increasingly large volume of cached states. This memory pressure severely limits the number of concurrent users a single accelerator can serve, as KV cache consumption scales linearly with both sequence length and batch size. \textbf{\textsc{Glide}} mitigates this by reducing aggregate KV retrieval through layer-wise hybrid attention. Non-uniform configurations bring KV I/O down to 5.5-11.0 GB at 128k tokens, a reduction of up to 3.2$\times$ compared to baseline. This reduction directly enables higher serving capacity: by compressing per-request memory footprint without major performance degradation (retaining 92-96\% accuracy), \textbf{\textsc{Glide}} allows deployments to serve 2-3$\times$ more concurrent users on the same hardware. \textit{The efficiency gains compound at scale, as reduced memory bandwidth contention improves overall throughput and reduces tail latency under load, making long-context inference practically deployable in multi-user serving environments.}

\section{Limitations \& Future Works}

While \textbf{\textsc{Glide}} demonstrates substantial improvements in decoding efficiency through selective KV cache retrieval and layer-wise hybrid attention, several limitations remain. Our experiments reveal that linear attention components exhibit information dilution as sequence length increases, where the compressed recurrent state progressively loses fine-grained token-level distinctions. Although the sliding-window softmax layers partially compensate by preserving local precision, this dilution contributes to the observed accuracy-efficiency trade-off on long-context workloads. Investigating improved state compression mechanisms or hybrid state management strategies that mitigate dilution while preserving memory efficiency remains an important direction. From a systems perspective, the current implementation relies on attention kernels not specifically optimized for hybrid execution patterns. We observed that smaller window sizes ($w \leq 128$), despite reducing theoretical FLOPs, incur kernel launch and memory scheduling overhead that diminishes practical latency gains on GPU architectures. Fused kernels co-optimized for sliding-window softmax and linear recurrent components, potentially leveraging persistent thread blocks or warp-specialized execution, could unlock further performance improvements. Finally, extending \textbf{\textsc{Glide}} to distributed inference scenarios with tensor or pipeline parallelism, and evaluating its impact on reasoning-intensive workloads where layer-wise attention sensitivity may differ, represent promising avenues for future work.

\section{Conclusion}
This paper introduces \textbf{\textsc{Glide}} (Guided Layerwise Hybrid Attention for Efficient Decoding), an adaptive framework that enables efficient long-context generation in large language models while preserving downstream task accuracy. \textbf{\textsc{Glide}} is built on a key empirical observation: transformer layers exhibit depth-dependent sensitivity to attention linearization, with early layers requiring high-fidelity softmax attention for representational quality, while deeper layers tolerate aggressive linearization with minimal performance impact. Leveraging this insight, \textbf{\textsc{Glide}} selectively allocates softmax attention non-uniformly across layers, concentrating expensive full-attention computation where most critical while replacing approximation-tolerant layers with memory-efficient linear recurrence. This layer-wise allocation strategy substantially reduces KV cache I/O during autoregressive decoding, the primary bottleneck in long-context generation, achieving up to 62$\times$ reduction in memory bandwidth while retaining 92-96\% of baseline accuracy across reasoning benchmarks. Beyond single-request efficiency, \textbf{\textsc{Glide}}'s compressed memory footprint directly translates to improved serving capacity: by reducing per-request KV cache overhead, the framework enables resource-constrained accelerators to handle 2-3$\times$ more concurrent users compared to vanilla softmax attention, making scalable multi-user deployment of long-context models practically feasible in memory-limited environments.

\bibliographystyle{IEEEtran}  
\bibliography{glide}

@article{vaswani2017attention,
  title={Attention is all you need},
  author={Vaswani, Ashish and Shazeer, Noam and Parmar, Niki and Uszkoreit, Jakob and Jones, Llion and Gomez, Aidan N and Kaiser, {\L}ukasz and Polosukhin, Illia},
  journal={Advances in neural information processing systems},
  volume={30},
  year={2017},
  url = {https://doi.org/10.48550/arXiv.1706.03762}
}

@article{beltagy2020longformer,
  title={Longformer: The long-document transformer},
  author={Beltagy, Iz and Peters, Matthew E and Cohan, Arman},
  journal={arXiv preprint arXiv:2004.05150},
  year={2020}, 
  url = {https://doi.org/10.48550/arXiv.2004.05150}
}

@article{arora2024simple,
  title={Simple linear attention language models balance the recall-throughput tradeoff},
  author={Arora, Simran and Eyuboglu, Sabri and Zhang, Michael and Timalsina, Aman and Alberti, Silas and Zinsley, Dylan and Zou, James and Rudra, Atri and R{\'e}, Christopher},
  journal={arXiv preprint arXiv:2402.18668},
  year={2024}, 
  url = {https://doi.org/10.48550/arXiv.2402.18668}
}

@article{cai2024pyramidkv,
  title={Pyramidkv: Dynamic kv cache compression based on pyramidal information funneling},
  author={Cai, Zefan and Zhang, Yichi and Gao, Bofei and Liu, Yuliang and Li, Yucheng and Liu, Tianyu and Lu, Keming and Xiong, Wayne and Dong, Yue and Hu, Junjie and others},
  journal={arXiv preprint arXiv:2406.02069},
  year={2024}, 
  url = {https://doi.org/10.48550/arXiv.2406.02069}
}

@article{li2024snapkv,
  title={Snapkv: Llm knows what you are looking for before generation},
  author={Li, Yuhong and Huang, Yingbing and Yang, Bowen and Venkitesh, Bharat and Locatelli, Acyr and Ye, Hanchen and Cai, Tianle and Lewis, Patrick and Chen, Deming},
  journal={Advances in Neural Information Processing Systems},
  volume={37},
  pages={22947--22970},
  year={2024},
  url = {https://doi.org/10.48550/arXiv.2404.14469}
}

@article{xiao2024efficient,
  title={Efficient streaming language models with attention sinks, 2024},
  author={Xiao, Guangxuan and Tian, Yuandong and Chen, Beidi and Han, Song and Lewis, Mike},
  journal={URL https://arxiv. org/abs/2309.17453},
  volume={1},
  year={2024}, 
  url = {https://doi.org/10.48550/arXiv.2309.17453}
}

@article{feng2024ada,
  title={Ada-kv: Optimizing kv cache eviction by adaptive budget allocation for efficient llm inference},
  author={Feng, Yuan and Lv, Junlin and Cao, Yukun and Xie, Xike and Zhou, S Kevin},
  journal={arXiv preprint arXiv:2407.11550},
  year={2024}, 
  url = {https://doi.org/10.48550/arXiv.2407.11550}
  
}

@article{zhang2023h2o,
  title={H2o: Heavy-hitter oracle for efficient generative inference of large language models},
  author={Zhang, Zhenyu and Sheng, Ying and Zhou, Tianyi and Chen, Tianlong and Zheng, Lianmin and Cai, Ruisi and Song, Zhao and Tian, Yuandong and R{\'e}, Christopher and Barrett, Clark and others},
  journal={Advances in Neural Information Processing Systems},
  volume={36},
  pages={34661--34710},
  year={2023}, 
  url = {https://doi.org/10.48550/arXiv.2306.14048}
}

@article{dong2024get,
  title={Get more with less: Synthesizing recurrence with kv cache compression for efficient llm inference},
  author={Dong, Harry and Yang, Xinyu and Zhang, Zhenyu and Wang, Zhangyang and Chi, Yuejie and Chen, Beidi},
  journal={arXiv preprint arXiv:2402.09398},
  year={2024}, 
  url = {https://doi.org/10.48550/arXiv.2402.09398}
}

@article{bridging_han_2024,
	title = {Bridging the Divide: Reconsidering Softmax and Linear Attention},
	doi = {10.48550/arxiv.2412.06590},
	author = {Han, Dongchen and Pu, Yifan and Xia, Zhuofan and Han, Yizeng and Pan, Xuran and Li, Xiu and Lu, Jiwen and Song, Shiji and Huang, Gao},
	journal = {arXiv.org},
	year = {2024},
	litmapsId = {282111368}, 
    url = {https://doi.org/10.48550/arXiv.2412.06590}
}

@article{xu2025mswa,
  title={MSWA: Refining Local Attention with Multi-ScaleWindow Attention},
  author={Xu, Yixing and Nag, Shivank and Li, Dong and Tian, Lu and Barsoum, Emad},
  journal={arXiv preprint arXiv:2501.01039},
  year={2025}, 
  url = {https://doi.org/10.48550/arXiv.2501.01039}
}

@article{munkhdalai2024leave,
  title={Leave no context behind: Efficient infinite context transformers with infini-attention},
  author={Munkhdalai, Tsendsuren and Faruqui, Manaal and Gopal, Siddharth},
  journal={arXiv preprint arXiv:2404.07143},
  volume={101},
  year={2024}, 
  url = {https://doi.org/10.48550/arXiv.2404.07143}
}

@article{chen2025edgeinfinite,
  title={Edgeinfinite: A memory-efficient infinite-context transformer for edge devices},
  author={Chen, Jiyu and Peng, Shuang and Luo, Daxiong and Yang, Fan and Wu, Renshou and Li, Fangyuan and Chen, Xiaoxin},
  journal={arXiv preprint arXiv:2503.22196},
  year={2025}, 
  url = {https://doi.org/10.48550/arXiv.2503.22196}
}

@inproceedings{dass2023vitality,
  title={Vitality: Unifying low-rank and sparse approximation for vision transformer acceleration with a linear taylor attention},
  author={Dass, Jyotikrishna and Wu, Shang and Shi, Huihong and Li, Chaojian and Ye, Zhifan and Wang, Zhongfeng and Lin, Yingyan},
  booktitle={2023 IEEE International Symposium on High-Performance Computer Architecture (HPCA)},
  pages={415--428},
  year={2023},
  organization={IEEE}, 
  url = {https://doi.org/10.48550/arXiv.2211.05109}
}

@article{xu2024qt,
  title={QT-ViT: Improving Linear Attention in ViT with Quadratic Taylor Expansion},
  author={Xu, Yixing and Li, Chao and Li, Dong and Sheng, Xiao and Jiang, Fan and Tian, Lu and Barsoum, Emad},
  journal={Advances in Neural Information Processing Systems},
  volume={37},
  pages={83048--83067},
  year={2024}, 
  url = {https://doi.org/10.52202/079017-2642}
}

@inproceedings{katharopoulos2020transformers,
  title={Transformers are rnns: Fast autoregressive transformers with linear attention},
  author={Katharopoulos, Angelos and Vyas, Apoorv and Pappas, Nikolaos and Fleuret, Fran{\c{c}}ois},
  booktitle={International conference on machine learning},
  pages={5156--5165},
  year={2020},
  organization={PMLR}, 
  url = {https://doi.org/10.48550/arXiv.2006.16236}
}

@article{han2024bridging,
  title={Bridging the divide: Reconsidering softmax and linear attention},
  author={Han, Dongchen and Pu, Yifan and Xia, Zhuofan and Han, Yizeng and Pan, Xuran and Li, Xiu and Lu, Jiwen and Song, Shiji and Huang, Gao},
  journal={Advances in Neural Information Processing Systems},
  volume={37},
  pages={79221--79245},
  year={2024}, 
  url = {https://doi.org/10.48550/arXiv.2412.06590}
}

@article{zhang2402hedgehog,
  title={The hedgehog \& the porcupine: Expressive linear attentions with softmax mimicry, 2024b},
  author={Zhang, Michael and Bhatia, Kush and Kumbong, Hermann and R{\'e}, Christopher},
  journal={URL https://arxiv. org/abs/2402.04347},
  year={2024}, 
  url = {https://doi.org/10.48550/arXiv.2402.04347}
}

@inproceedings{
zhang2025lolcats,
title={Lo{LCAT}s: On Low-Rank Linearizing of Large Language Models},
author={Michael Zhang and Simran Arora and Rahul Chalamala and Benjamin Frederick Spector and Alan Wu and Krithik Ramesh and Aaryan Singhal and Christopher Re},
booktitle={The Thirteenth International Conference on Learning Representations},
year={2025},
url={https://doi.org/10.48550/arXiv.2410.10254}
}

@inproceedings{
hu2022lora,
title={Lo{RA}: Low-Rank Adaptation of Large Language Models},
author={Edward J Hu and yelong shen and Phillip Wallis and Zeyuan Allen-Zhu and Yuanzhi Li and Shean Wang and Lu Wang and Weizhu Chen},
booktitle={International Conference on Learning Representations},
year={2022},
url={https://doi.org/10.48550/arXiv.2106.09685}
}

@inproceedings{
lan2025liger,
title={Liger: Linearizing Large Language Models to Gated Recurrent Structures},
author={Disen Lan and Weigao Sun and Jiaxi Hu and Jusen Du and Yu Cheng},
booktitle={Forty-second International Conference on Machine Learning},
year={2025},
url={https://doi.org/10.48550/arXiv.2503.01496}
}

@inproceedings{
dong2025flexattention,
title={FlexAttention: A Programming Model for Generating Fused Attention Variants.},
author={Juechu Dong and BOYUAN FENG and Driss Guessous and Yanbo Liang and Horace He},
booktitle={Eighth Conference on Machine Learning and Systems},
year={2025},
url={https://doi.org/10.48550/arXiv.2412.05496}
}

@misc{alpaca,
  author = {Rohan Taori and Ishaan Gulrajani and Tianyi Zhang and Yann Dubois and Xuechen Li and Carlos Guestrin and Percy Liang and Tatsunori B. Hashimoto },
  title = {Stanford Alpaca: An Instruction-following LLaMA model},
  year = {2023},
  publisher = {GitHub},
  journal = {GitHub repository},
  howpublished = {\url{https://github.com/tatsu-lab/stanford_alpaca}},
}

@misc{grattafiori2024llama3herdmodels,
      title={The Llama 3 Herd of Models}, 
      author={Aaron Grattafiori and Abhimanyu Dubey and Abhinav Jauhri and Abhinav Pandey and Abhishek Kadian and Ahmad Al-Dahle and Aiesha Letman and Akhil Mathur and Alan Schelten and Alex Vaughan and Amy Yang and Angela Fan and Anirudh Goyal and Anthony Hartshorn and Aobo Yang and Archi Mitra and Archie Sravankumar and Artem Korenev and Arthur Hinsvark and Arun Rao and Aston Zhang and Aurelien Rodriguez and Austen Gregerson and Ava Spataru and Baptiste Roziere and Bethany Biron and Binh Tang and Bobbie Chern and Charlotte Caucheteux and Chaya Nayak and Chloe Bi and Chris Marra and Chris McConnell and Christian Keller and Christophe Touret and Chunyang Wu and Corinne Wong and Cristian Canton Ferrer and Cyrus Nikolaidis and Damien Allonsius and Daniel Song and Danielle Pintz and Danny Livshits and Danny Wyatt and David Esiobu and Dhruv Choudhary and Dhruv Mahajan and Diego Garcia-Olano and Diego Perino and Dieuwke Hupkes and Egor Lakomkin and Ehab AlBadawy and Elina Lobanova and Emily Dinan and Eric Michael Smith and Filip Radenovic and Francisco Guzmán and Frank Zhang and Gabriel Synnaeve and Gabrielle Lee and Georgia Lewis Anderson and Govind Thattai and Graeme Nail and Gregoire Mialon and Guan Pang and Guillem Cucurell and Hailey Nguyen and Hannah Korevaar and Hu Xu and Hugo Touvron and Iliyan Zarov and Imanol Arrieta Ibarra and Isabel Kloumann and Ishan Misra and Ivan Evtimov and Jack Zhang and Jade Copet and Jaewon Lee and Jan Geffert and Jana Vranes and Jason Park and Jay Mahadeokar and Jeet Shah and Jelmer van der Linde and Jennifer Billock and Jenny Hong and Jenya Lee and Jeremy Fu and Jianfeng Chi and Jianyu Huang and Jiawen Liu and Jie Wang and Jiecao Yu and Joanna Bitton and Joe Spisak and Jongsoo Park and Joseph Rocca and Joshua Johnstun and Joshua Saxe and Junteng Jia and Kalyan Vasuden Alwala and Karthik Prasad and Kartikeya Upasani and Kate Plawiak and Ke Li and Kenneth Heafield and Kevin Stone and Khalid El-Arini and Krithika Iyer and Kshitiz Malik and Kuenley Chiu and Kunal Bhalla and Kushal Lakhotia and Lauren Rantala-Yeary and Laurens van der Maaten and Lawrence Chen and Liang Tan and Liz Jenkins and Louis Martin and Lovish Madaan and Lubo Malo and Lukas Blecher and Lukas Landzaat and Luke de Oliveira and Madeline Muzzi and Mahesh Pasupuleti and Mannat Singh and Manohar Paluri and Marcin Kardas and Maria Tsimpoukelli and Mathew Oldham and Mathieu Rita and Maya Pavlova and Melanie Kambadur and Mike Lewis and Min Si and Mitesh Kumar Singh and Mona Hassan and Naman Goyal and Narjes Torabi and Nikolay Bashlykov and Nikolay Bogoychev and Niladri Chatterji and Ning Zhang and Olivier Duchenne and Onur Çelebi and Patrick Alrassy and Pengchuan Zhang and Pengwei Li and Petar Vasic and Peter Weng and Prajjwal Bhargava and Pratik Dubal and Praveen Krishnan and Punit Singh Koura and Puxin Xu and Qing He and Qingxiao Dong and Ragavan Srinivasan and Raj Ganapathy and Ramon Calderer and Ricardo Silveira Cabral and Robert Stojnic and Roberta Raileanu and Rohan Maheswari and Rohit Girdhar and Rohit Patel and Romain Sauvestre and Ronnie Polidoro and Roshan Sumbaly and Ross Taylor and Ruan Silva and Rui Hou and Rui Wang and Saghar Hosseini and Sahana Chennabasappa and Sanjay Singh and Sean Bell and Seohyun Sonia Kim and Sergey Edunov and Shaoliang Nie and Sharan Narang and Sharath Raparthy and Sheng Shen and Shengye Wan and Shruti Bhosale and Shun Zhang and Simon Vandenhende and Soumya Batra and Spencer Whitman and Sten Sootla and Stephane Collot and Suchin Gururangan and Sydney Borodinsky and Tamar Herman and Tara Fowler and Tarek Sheasha and Thomas Georgiou and Thomas Scialom and Tobias Speckbacher and Todor Mihaylov and Tong Xiao and Ujjwal Karn and Vedanuj Goswami and Vibhor Gupta and Vignesh Ramanathan and Viktor Kerkez and Vincent Gonguet and Virginie Do and Vish Vogeti and Vítor Albiero and Vladan Petrovic and Weiwei Chu and Wenhan Xiong and Wenyin Fu and Whitney Meers and Xavier Martinet and Xiaodong Wang and Xiaofang Wang and Xiaoqing Ellen Tan and Xide Xia and Xinfeng Xie and Xuchao Jia and Xuewei Wang and Yaelle Goldschlag and Yashesh Gaur and Yasmine Babaei and Yi Wen and Yiwen Song and Yuchen Zhang and Yue Li and Yuning Mao and Zacharie Delpierre Coudert and Zheng Yan and Zhengxing Chen and Zoe Papakipos and Aaditya Singh and Aayushi Srivastava and Abha Jain and Adam Kelsey and Adam Shajnfeld and Adithya Gangidi and Adolfo Victoria and Ahuva Goldstand and Ajay Menon and Ajay Sharma and Alex Boesenberg and Alexei Baevski and Allie Feinstein and Amanda Kallet and Amit Sangani and Amos Teo and Anam Yunus and Andrei Lupu and Andres Alvarado and Andrew Caples and Andrew Gu and Andrew Ho and Andrew Poulton and Andrew Ryan and Ankit Ramchandani and Annie Dong and Annie Franco and Anuj Goyal and Aparajita Saraf and Arkabandhu Chowdhury and Ashley Gabriel and Ashwin Bharambe and Assaf Eisenman and Azadeh Yazdan and Beau James and Ben Maurer and Benjamin Leonhardi and Bernie Huang and Beth Loyd and Beto De Paola and Bhargavi Paranjape and Bing Liu and Bo Wu and Boyu Ni and Braden Hancock and Bram Wasti and Brandon Spence and Brani Stojkovic and Brian Gamido and Britt Montalvo and Carl Parker and Carly Burton and Catalina Mejia and Ce Liu and Changhan Wang and Changkyu Kim and Chao Zhou and Chester Hu and Ching-Hsiang Chu and Chris Cai and Chris Tindal and Christoph Feichtenhofer and Cynthia Gao and Damon Civin and Dana Beaty and Daniel Kreymer and Daniel Li and David Adkins and David Xu and Davide Testuggine and Delia David and Devi Parikh and Diana Liskovich and Didem Foss and Dingkang Wang and Duc Le and Dustin Holland and Edward Dowling and Eissa Jamil and Elaine Montgomery and Eleonora Presani and Emily Hahn and Emily Wood and Eric-Tuan Le and Erik Brinkman and Esteban Arcaute and Evan Dunbar and Evan Smothers and Fei Sun and Felix Kreuk and Feng Tian and Filippos Kokkinos and Firat Ozgenel and Francesco Caggioni and Frank Kanayet and Frank Seide and Gabriela Medina Florez and Gabriella Schwarz and Gada Badeer and Georgia Swee and Gil Halpern and Grant Herman and Grigory Sizov and Guangyi and Zhang and Guna Lakshminarayanan and Hakan Inan and Hamid Shojanazeri and Han Zou and Hannah Wang and Hanwen Zha and Haroun Habeeb and Harrison Rudolph and Helen Suk and Henry Aspegren and Hunter Goldman and Hongyuan Zhan and Ibrahim Damlaj and Igor Molybog and Igor Tufanov and Ilias Leontiadis and Irina-Elena Veliche and Itai Gat and Jake Weissman and James Geboski and James Kohli and Janice Lam and Japhet Asher and Jean-Baptiste Gaya and Jeff Marcus and Jeff Tang and Jennifer Chan and Jenny Zhen and Jeremy Reizenstein and Jeremy Teboul and Jessica Zhong and Jian Jin and Jingyi Yang and Joe Cummings and Jon Carvill and Jon Shepard and Jonathan McPhie and Jonathan Torres and Josh Ginsburg and Junjie Wang and Kai Wu and Kam Hou U and Karan Saxena and Kartikay Khandelwal and Katayoun Zand and Kathy Matosich and Kaushik Veeraraghavan and Kelly Michelena and Keqian Li and Kiran Jagadeesh and Kun Huang and Kunal Chawla and Kyle Huang and Lailin Chen and Lakshya Garg and Lavender A and Leandro Silva and Lee Bell and Lei Zhang and Liangpeng Guo and Licheng Yu and Liron Moshkovich and Luca Wehrstedt and Madian Khabsa and Manav Avalani and Manish Bhatt and Martynas Mankus and Matan Hasson and Matthew Lennie and Matthias Reso and Maxim Groshev and Maxim Naumov and Maya Lathi and Meghan Keneally and Miao Liu and Michael L. Seltzer and Michal Valko and Michelle Restrepo and Mihir Patel and Mik Vyatskov and Mikayel Samvelyan and Mike Clark and Mike Macey and Mike Wang and Miquel Jubert Hermoso and Mo Metanat and Mohammad Rastegari and Munish Bansal and Nandhini Santhanam and Natascha Parks and Natasha White and Navyata Bawa and Nayan Singhal and Nick Egebo and Nicolas Usunier and Nikhil Mehta and Nikolay Pavlovich Laptev and Ning Dong and Norman Cheng and Oleg Chernoguz and Olivia Hart and Omkar Salpekar and Ozlem Kalinli and Parkin Kent and Parth Parekh and Paul Saab and Pavan Balaji and Pedro Rittner and Philip Bontrager and Pierre Roux and Piotr Dollar and Polina Zvyagina and Prashant Ratanchandani and Pritish Yuvraj and Qian Liang and Rachad Alao and Rachel Rodriguez and Rafi Ayub and Raghotham Murthy and Raghu Nayani and Rahul Mitra and Rangaprabhu Parthasarathy and Raymond Li and Rebekkah Hogan and Robin Battey and Rocky Wang and Russ Howes and Ruty Rinott and Sachin Mehta and Sachin Siby and Sai Jayesh Bondu and Samyak Datta and Sara Chugh and Sara Hunt and Sargun Dhillon and Sasha Sidorov and Satadru Pan and Saurabh Mahajan and Saurabh Verma and Seiji Yamamoto and Sharadh Ramaswamy and Shaun Lindsay and Shaun Lindsay and Sheng Feng and Shenghao Lin and Shengxin Cindy Zha and Shishir Patil and Shiva Shankar and Shuqiang Zhang and Shuqiang Zhang and Sinong Wang and Sneha Agarwal and Soji Sajuyigbe and Soumith Chintala and Stephanie Max and Stephen Chen and Steve Kehoe and Steve Satterfield and Sudarshan Govindaprasad and Sumit Gupta and Summer Deng and Sungmin Cho and Sunny Virk and Suraj Subramanian and Sy Choudhury and Sydney Goldman and Tal Remez and Tamar Glaser and Tamara Best and Thilo Koehler and Thomas Robinson and Tianhe Li and Tianjun Zhang and Tim Matthews and Timothy Chou and Tzook Shaked and Varun Vontimitta and Victoria Ajayi and Victoria Montanez and Vijai Mohan and Vinay Satish Kumar and Vishal Mangla and Vlad Ionescu and Vlad Poenaru and Vlad Tiberiu Mihailescu and Vladimir Ivanov and Wei Li and Wenchen Wang and Wenwen Jiang and Wes Bouaziz and Will Constable and Xiaocheng Tang and Xiaojian Wu and Xiaolan Wang and Xilun Wu and Xinbo Gao and Yaniv Kleinman and Yanjun Chen and Ye Hu and Ye Jia and Ye Qi and Yenda Li and Yilin Zhang and Ying Zhang and Yossi Adi and Youngjin Nam and Yu and Wang and Yu Zhao and Yuchen Hao and Yundi Qian and Yunlu Li and Yuzi He and Zach Rait and Zachary DeVito and Zef Rosnbrick and Zhaoduo Wen and Zhenyu Yang and Zhiwei Zhao and Zhiyu Ma},
      year={2024},
      eprint={2407.21783},
      archivePrefix={arXiv},
      primaryClass={cs.AI},
      url={https://doi.org/10.48550/arXiv.2407.21783}
}

@misc{jiang2023mistral7b,
      title={Mistral 7B}, 
      author={Albert Q. Jiang and Alexandre Sablayrolles and Arthur Mensch and Chris Bamford and Devendra Singh Chaplot and Diego de las Casas and Florian Bressand and Gianna Lengyel and Guillaume Lample and Lucile Saulnier and Lélio Renard Lavaud and Marie-Anne Lachaux and Pierre Stock and Teven Le Scao and Thibaut Lavril and Thomas Wang and Timothée Lacroix and William El Sayed},
      year={2023},
      eprint={2310.06825},
      archivePrefix={arXiv},
      primaryClass={cs.CL},
      url={https://doi.org/10.48550/arXiv.2310.06825}
}

@article{
    li2025a,
    title={A Survey on Large Language Model Acceleration based on {KV} Cache Management},
    author={Haoyang LI and Yiming Li and Anxin Tian and Tianhao Tang and Zhanchao Xu and Xuejia Chen and Nicole HU and Wei Dong and Li Qing and Lei Chen},
    journal={Transactions on Machine Learning Research},
    issn={2835-8856},
    year={2025},
    url={https://doi.org/10.48550/arXiv.2412.19442},
    note={}
}

\clearpage
\appendix \label{sec:appd}

\textbf{Overview of Attention.} \label{sec: attention_overview} Consider the next token prediction using the current query $\mathbf{q}_i \in \mathbb{R}^{d}$, where $d = d_{\text{model}} / H$ and $H$ denote the number of heads present in the single multi-head attention layer and $d_{\text{model}}$ is the overall dimension of the model.  The attention score $\alpha_{ij}$ captures the similarity between the current query $\mathbf{q}_i$ and the keys $\mathbf{k}_j \in \mathbb{R}^{d}$, where $j \in [1,i]$ defined as, 

\begin{equation} \label{1}
    \alpha_{ij} = \frac{ \text{sim}(\mathbf{q}_i, \mathbf{k}_j)}{\sum_{j'=1}^{i} \text{sim}(\mathbf{q}_i, \mathbf{k}_{j'})},
\end{equation}

\begin{equation} \label{2}
    O_i = \sum_{j=1}^i \alpha_{ij} \mathbf{v}_j,
\end{equation}

where, $ \text{sim}(\mathbf{q}_i,\mathbf{k}_j)= \text{exp}(\mathbf{q}_i,\mathbf{k}_j^T/\sqrt{d})$ corresponds to softmax attention \cite{vaswani2017attention}. The output of the attention block $O_i$ defined in \eqref{1} is the attention-weighted aggregation of value vectors $\mathbf{v}_j \in \mathbb{R}^{d}$. Further, the computed output $O_i$ is transformed linearly by dot product with the projection matrix $W_o \in \mathbb{R}^{n \times d}$, where $n$ denotes the context length of the model. According to \eqref{1}, predicting $i+1^{th}$ token depends on the set of previously generated key \& value pairs $\{\, \mathbf{k}_j, \mathbf{v}_j \mid j \in [1,i] \,\}$ and computing those matrices at time $t$ could require expensive $O(n^2)$ computational cost.

\textbf{KV Cache.} \label{sec:kv_cache} The overhead involved in computing the past key and value pairs can be mitigated by caching each generated key $\mathbf{k_i}$ and $\mathbf{v_i}$ at every iteration. KV caching involves two computational stages namely prefill and decode. \textit{Prefill Stage:} The stage during which the model computes self-attention over all prompts, and populates the key–value (KV) cache by storing the key $\mathbf{k_i}$ and value $\mathbf{v_i}$ pairs for each token.\textit{ Decode Stage:} The stage in which the model generates tokens auto-regressively; at each step, self-attention is computed using the previously stored key–value (KV) cache, and the key and value corresponding to the newly generated token are appended to the cache. Despite minimizing redundant computes, the memory required for the KV cache scales linearly with the context length $n$, and at each token prediction, attention incurs an expensive $O(n)$ memory footprint.

\textbf{Linear Attention.} To understand how global context is preserved in retention-based SWA, we formalize linear attention, which underlies these methods. Considering \eqref{1}, the similarity function $sim(\cdot)$ in softmax attention can be approximated using separable kernel functions $\phi$~\cite{katharopoulos2020transformers}, applied independently to the query $\mathbf{q}_i$ and key $\mathbf{k}_j$. Under this formulation, the approximated attention score $\alpha_{ij}$ can be expressed as,

\begin{equation} \label{5}
    \frac{\text{sim}(\mathbf{q}_i.\mathbf{k}_j)}{\sum_{j=1}^i \text{sim}(\mathbf{q}_i.\mathbf{k}_j)} \approx \frac{\phi(\mathbf{q}_i).\phi(\mathbf{k}_j)^T}{\sum_{j=1}^i \phi(\mathbf{q}_i).\phi(\mathbf{k}_j)^T}, 
\end{equation}

on computing the $O_i$, the matrices can be rearranged from $\sum_{j=i}^i\phi(\mathbf{q}_i).\phi(\mathbf{k}_j)^T\mathbf{v}_j$ to $\phi(\mathbf{q}_i).(\sum_{j=i}^i\phi(\mathbf{k}_j)^T\mathbf{v}_j)$ resulting in the constant recurrent memory of $d \times d$ and it can be expressed as, 

\begin{equation} \label{6}
    O_i^{\text{Linear}} = \frac{\phi(\mathbf{q}_i).(\sum_{j=1}^i\phi(\mathbf{k}_j)^T.\mathbf{v}_j)}{\phi(\mathbf{q}_i).\sum_{j=1}^i \phi(\mathbf{k}_j)^T} = \frac{\phi(\mathbf{q}_i).\mathbf{S}_i}{\phi(\mathbf{q}_i).\mathbf{Z}_i},
\end{equation}

where the recurrent state $\mathbf{S}_i$ is updated as $\mathbf{S}_i = \mathbf{S}_{i-1} + \phi(\mathbf{k}_j)^{\top} \mathbf{v}_j \in \mathbb{R}^{d \times d} $, and the corresponding normalization factor is updated as $\mathbf{Z}_i = \mathbf{Z}_{i-1} + \sum_{j=1}^{i} \phi(\mathbf{k}_j)^{\top} \in \mathbb{R}^{d} $. Unlike softmax attention, linear attention incurs only an \(O(d)\) data movement cost, and its memory usage remains constant with respect to the context length \(n\). Vanilla softmax attention can be approximated using linear attention with different choices of \(\phi\)-kernels. Commonly used \(\phi\)-kernels include \(\mathrm{ELU}+1\) (EdgeInfinite~\cite{chen2025edgeinfinite}, Infini-Transformer~\cite{munkhdalai2024leave}), MLP-based kernels (LESS~\cite{dong2024get}, Hedgehog~\cite{zhang2402hedgehog}, LoLCats~\cite{zhang2025lolcats}, LOLA~\cite{hu2022lora}), ReLU (InLine~\cite{han2024bridging}), and Taylor-series expansions (ViTaLity~\cite{dass2023vitality}(degree-1), QT-ViT~\cite{xu2024qt}(degree-2), BASED~\cite{arora2024simple}(degree~2)) and Softmax (Liger~\cite{lan2025liger}).

\subsection{Extended Study on Window Attention} \label{sec:swa}
\textbf{Sliding Window Attention.} To alleviate the linear growth in memory footprint and per-token attention cost induced by full KV caching, prior work proposes Sliding Window Attention \cite{beltagy2020longformer}. Under sliding window attention, self-attention for query $\mathbf{q}_i$ is restricted to the most recent window size $w$ key--value pairs, and the attention score $\alpha_{ij}$ is computed only within this window as,


\begin{equation} \label{3}
    \alpha_{ij}^{\text{SWA}} = \frac{ \text{exp}(\mathbf{q}_i, \mathbf{k}_j)}{\sum_{j'=i-w+1}^{i} \text{exp}(\mathbf{q}_i, \mathbf{k}_{j'})},
\end{equation}

\begin{equation} \label{4}
    O_{i}^{\text{SWA}} = \sum_{j=i-w+1}^i \alpha^{\text{SWA}}_{ij} . \mathbf{v}_j,
\end{equation}

where the attention incurs a reduced $O(w)$ memory footprint, with the window size 
$w$ constrained such that $w < n$. In SWA, based on the window size $w$ the initial token within the window gets evicted as the new token comes in, this first-in based eviction policy could remove important tokens from the window, leading to a degradation in the model’s recall rate, which measures the model’s ability to retrieve, attend to, and condition its generation on relevant information from earlier context tokens.

\noindent \textbf{(a) Eviction-based SWA.}
 Eviction-based approaches compute attention scores under two categories: a fixed KV cache and an adaptive KV cache budget. Some of the fixed-cache budget works are SnapKV~\cite{li2024snapkv} compresses the KV cache by retaining head-specific salient tokens identified from attention patterns. H$_2$O~\cite{zhang2023h2o} dynamically manages the KV cache by preserving both recent tokens and heavy hitters with high cumulative attention scores. StreamingLLM~\cite{xiao2024efficient} enables unbounded sequence generation by explicitly retaining attention sink tokens. Beyond fixed-budget eviction, several works introduce adaptive KV-cache budget allocation strategies. PyramidKV~\cite{cai2024pyramidkv} proposes dynamic KV-cache memory allocation across transformer layers, selectively retaining important KV pairs under a global budget constraint.

 \begin{figure*}[t]
    \centering

        \includegraphics[width=0.85\textwidth]{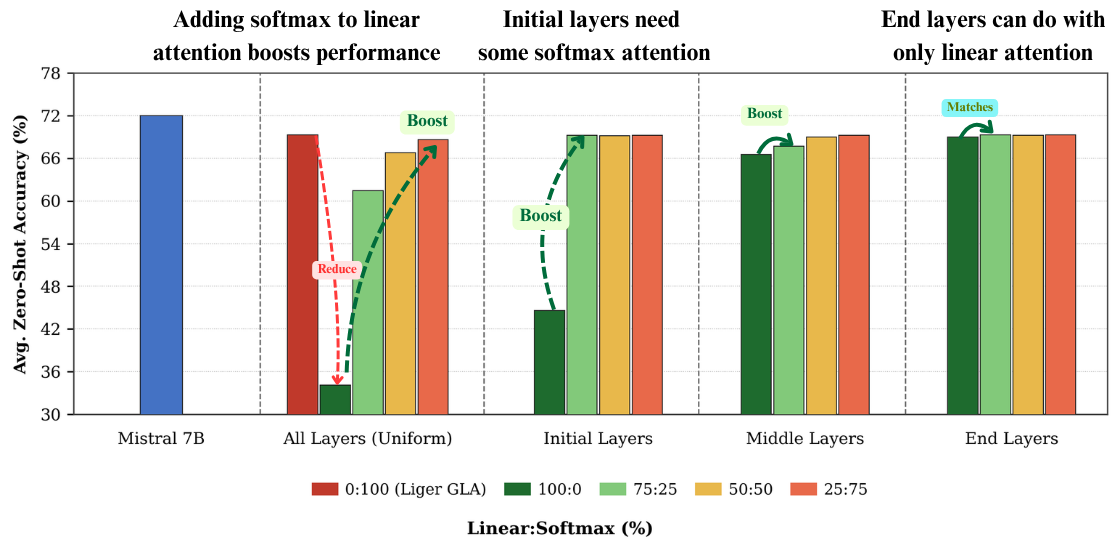}%
        \label{fig:layer_sensitivity_mist}%
    \caption{\textbf{Analysis of Zero-Shot Accuracy Under Varying Softmax Attention Intensity (Non~Fine-Tuned).} Layer-wise sensitivity analysis: initial layers require softmax attention, while end layers tolerate full linearization. We evaluate uniform configurations $\{0,\, w/4,\, w/2,\, 3w/4,\, w\}$ with the base window size $w=64$ and model dimension $d=4096$ across all transformer layers using LM-Eval (PiQA, ARC-e, ARC-c, HellaSwag, WinoG), then examine recovery by applying varying softmax attention to isolated layer groups (Early, Middle, Late).}
    \label{fig:abs_adx_1}
\end{figure*}

\noindent \textbf{(b) Retention-based SWA.}
Despite advances in local context optimization, eviction-based methods inherently restrict access to global context, which can adversely affect recall. This limitation can be mitigated by preserving global context as a recurrent state via linear attention mechanisms, enabling long-range information retention under a constant memory budget.  Building on this framework, LOLA~\cite{hu2022lora} introduces a sparse caching mechanism to reduce computational overhead while maintaining approximation quality. Additionally, the prior works formulates attention hybridization as a two-component output combining a linear recurrent with a local softmax term over the active window. The various mathematical formation of hybrid attention can be given by,

\begin{equation}
O^{\text{hybrid}}_i = 
\frac{\phi(\mathbf{q}_i)\,\mathbf{S}_{i-w}}
     {\phi(\mathbf{q}_i)\,\mathbf{Z}_{i-w}} 
\;+\; 
\frac{\sum_{j=i-w+1}^{i} \exp(\mathbf{q}_i \mathbf{k}_j^\top)\,\mathbf{v}_j}
     {\sum_{j'=i-w+1}^{i} \exp(\mathbf{q}_i \mathbf{k}_{j'}^\top)}
\label{eq:hybrid_v1}
\end{equation}

\noindent where $\mathbf{S}_{i-w} = \sum_{j=1}^{i-w} \phi(\mathbf{k}_j)^\top \mathbf{v}_j$ and $\mathbf{Z}_{i-w} = \sum_{j=1}^{i-w} \phi(\mathbf{k}_j)^\top$ are the recurrent state and normalization factor accumulating all tokens evicted from the sliding window. Further, the another form of hybrid attention can be written as,

\begin{equation}
O^{\text{hybrid}}_i = 
\frac{\phi(\mathbf{q}_i)\,\mathbf{S}_{i}}
     {\phi(\mathbf{q}_i)\,\mathbf{Z}_{i}} 
\;+\; 
\frac{\sum_{j=i-w+1}^{i} \exp(\mathbf{q}_i \mathbf{k}_j^\top)\,\mathbf{v}_j}
     {\sum_{j'=i-w+1}^{i} \exp(\mathbf{q}_i \mathbf{k}_{j'}^\top)}
\label{eq:hybrid_v2}
\end{equation}

\noindent where, the constant recurrent state $S_i \in \mathbb{R}^{d \times d}$ includes redundant information from the prefill or the instance before the sliding window attention starts evicting the tokens. \textbf{\textsc{Glide}} adapts both forms of the hybrid attention 
mechanism, unified through the generic $\oplus$ operator.

\end{document}